\documentclass[10pt,journal,compsoc]{IEEEtran}

\usepackage[utf8]{inputenc} 
\usepackage[T1]{fontenc}    
\usepackage{hyperref}       
\usepackage{url}            
\usepackage{booktabs}       
\usepackage{amsfonts}       
\usepackage{dsfont}
\usepackage{multirow}

\usepackage{nicefrac}       
\usepackage{microtype}      

\usepackage{amssymb}
\usepackage{subfigure}
\usepackage{multicol}
\usepackage{soul}

\usepackage{graphicx}
\usepackage{float}
\usepackage{amsmath}
\usepackage{bm}
\usepackage[misc]{ifsym}
\usepackage{pifont}
\newcommand{\cmark}{\ding{51}}%
\newcommand{\xmark}{\ding{55}}%
\usepackage{threeparttable}
\usepackage{multirow}
\usepackage{multicol}
\usepackage{color,colortbl}

\usepackage{tcolorbox}

\definecolor{my_blue}{rgb}{0.8431, 0.9300, 0.8431}
\definecolor{my_b}{rgb}{0.906, 0.945, 0.976}
\definecolor{my_g}{rgb}{0.898, 1.000, 0.898}
\definecolor{my_y}{rgb}{1.000, 0.973, 0.898}
\definecolor{my_darkgreen}{RGB}{30,150,30}
\definecolor{my_gray}{rgb}{0.925, 0.925, 0.925}
\newcolumntype{a}{>{\columncolor{my_gray}}c}

\makeatletter
\g@addto@macro{\endtabular}{\rowfont{}}
\makeatother
\newcommand{\rowfonttype}{}
\newcommand{\rowfont}[1]{
   \gdef\rowfonttype{#1}#1%
}
\newcolumntype{L}{>{\rowfonttype}l}

\newcolumntype{b}{>{\columncolor{my_b}}c}
\newcolumntype{g}{>{\columncolor{my_g}}c}
\newcolumntype{y}{>{\columncolor{my_y}}c}

\definecolor{my_r}{rgb}{0.949, 0.474, 0.439}

\DeclareMathAlphabet{\mymathbb}{U}{BOONDOX-ds}{m}{n}

\usepackage{ragged2e}

\hypersetup{
    colorlinks=true,
    linkcolor=black,
    filecolor=black,
    urlcolor=black,
    citecolor=black,
}

\begin{document}

\title{GLC++: Source-Free Universal Domain Adaptation through Global-Local Clustering and Contrastive Affinity Learning}
\author{Sanqing Qu, Tianpei Zou, Florian Röhrbein, Cewu Lu,\\
        Guang Chen{\textsuperscript{(\Letter)}}, Dacheng Tao,~\IEEEmembership{Fellow, IEEE},
        and Changjun Jiang
\IEEEcompsocitemizethanks{
\IEEEcompsocthanksitem Sanqing Qu, and Guang Chen are with the School of Computer Science and Technology, Tongji University, China. Email: \{sanqingqu, guangchen\}@tongji.edu.cn.
\IEEEcompsocthanksitem Tianpei Zou is with the School of Automotive Engineering, Tongji University, China. Email: 2011459@tongji.edu.cn.
\IEEEcompsocthanksitem Florian Röhrbein is with the Faculty of Computer Science, Chemnitz University of Technology, Germany. Email: florian.roehrbein@informatik.tu-chemnitz.de.
\IEEEcompsocthanksitem Cewu Lu is with the Department of Computer Science, Shanghai Jiao Tong University, China. Email: lucewu@sjtu.edu.cn.
\IEEEcompsocthanksitem Dacheng Tao is with the College of Computing and Data Science, Nanyang Technological University, Singapore. Email: dacheng.tao@gmail.com
\IEEEcompsocthanksitem Changjun Jiang is with the School of Computer Science and Technology, Key Laboratory of Embedded System and Service Computing, Ministry of Education, Tongji University, China. Email: cjjiang@tongji.edu.cn.
}
}

\markboth{Journal of \LaTeX\ Class Files, Feb~2024}%
{Shell \MakeLowercase{\textit{et al.}}: Source-Free Universal Domain Adaptation through Global and Local Clustering}

\IEEEtitleabstractindextext{%
\begin{abstract}
\justifying{
Deep neural networks often exhibit sub-optimal performance under covariate and category shifts. Source-Free Domain Adaptation (SFDA) presents a promising solution to this dilemma, yet most SFDA approaches are restricted to closed-set scenarios. In this paper, we explore Source-Free Universal Domain Adaptation (SF-UniDA) aiming to accurately classify "known" data belonging to common categories and segregate them from target-private "unknown" data. We propose a novel Global and Local Clustering (GLC) technique, which comprises an adaptive one-vs-all global clustering algorithm to discern between target classes, complemented by a local k-NN clustering strategy to mitigate negative transfer. Despite the effectiveness, the inherent closed-set source architecture leads to uniform treatment of "unknown" data, impeding the identification of distinct "unknown" categories. To address this, we evolve GLC to GLC++, integrating a contrastive affinity learning strategy. We examine the superiority of GLC and GLC++ across multiple benchmarks and category shift scenarios. Remarkably, in the most challenging open-partial-set scenarios, GLC and GLC++ surpass GATE by 16.8\% and 18.9\% in H-score on VisDA, respectively. GLC++ enhances the novel category clustering accuracy of GLC by 4.1\% in open-set scenarios on Office-Home. Furthermore, the introduced contrastive learning strategy not only enhances GLC but also significantly facilitates existing methodologies. The code is available at \url{https://github.com/ispc-lab/GLC-plus}.
}
\end{abstract}

\begin{IEEEkeywords}
Universal domain adaptation, source-free domain adaptation, clustering, contrastive learning, novel category discovery.
\end{IEEEkeywords}}

\maketitle\IEEEdisplaynontitleabstractindextext
\IEEEpeerreviewmaketitle

\section{Introduction}
\label{sec:intro}
\par \IEEEPARstart{A}{t} the expensive cost of given large-scale labeled data and huge computation resources, deep neural networks (DNNs) have demonstrated remarkable progress across a spectrum of tasks. However, these networks often struggle to generalize effectively to new, unseen domains, particularly in the presence of covariate and category shifts. How to upcycle DNNs and adapt them to specific target tasks is still a long-standing open problem. Over the past decade, many efforts have been devoted to unsupervised domain adaptation (UDA)~\cite{MMD, dann, hoffman2018cycada, MCD}. It capitalizes on labeled source data alongside unlabeled target data in a transductive learning manner, and has achieved significant success. Despite this, the access to source raw data is inefficient and may violate the increasingly stringent data privacy policies~\cite{GDPR_EU}. Recently, Source-Free Domain Adaptation (SFDA)~\cite{shot, gsfda, BMD} has become a promising technology to alleviate this issue, where only a pre-trained source model is provided as supervision rather than raw data. 
Despite its advantages, a prevalent limitation among most existing SFDA methods~\cite{shot, gsfda, BMD} is the presupposition of a consistent label space between the source and target domains. 

\par In practical applications, target data may originate from diverse situations, making the maintenance of such a strict assumption challenging and often unrealistic. For a better illustration, we suppose $\mathcal{Y}_s$ and $\mathcal{Y}_t$ as the label space of source domain and target domain, respectively. In addition to the well-studied closed-set domain adaptation (CLDA) where $\mathcal{Y}_s = \mathcal{Y}_t$, several other scenarios are prevalent. These include the partial-set domain adaptation (PDA)~\cite{pda, cao2018partial}, where the source domain encompasses a broader label space
$\mathcal{Y}_s \supset \mathcal{Y}_t$; the open-set domain adaptation (OSDA)~\cite{panareda2017open, saito2018open} where the target domain contains private, novel categories not present in the source domain ($\mathcal{Y}_s \subset \mathcal{Y}_t$); and the open-partial-set domain adaptation (OPDA)~\cite{uan, ova}, which involves a complex interaction of label spaces between source and target domain ($\mathcal{Y}_s \cap \mathcal{Y}_t \ne \emptyset$, $\mathcal{Y}_s \nsubseteq \mathcal{Y}_t$, $\mathcal{Y}_s \nsupseteq \mathcal{Y}_t$). To handle these generalized cases, universal domain adaptation (UniDA) has been proposed. The goal is to recognize target data belonging to source-target shared "known" categories and separate them from target-private "unknown" data. UniDA offers solutions~\cite{dance, dcc, gate} that accommodate all potential category shifts between source and target domains without prior knowledge of the target domain, e.g., information on matched common classes or the number of categories in the target domain.  Despite the progress, existing works~\cite{dance, dcc, gate} still necessitate concurrent access to source and target data, making it impractical and potentially violating data protection policies.

\begin{figure}[t]
    \centering
    \includegraphics[width=0.46\textwidth]{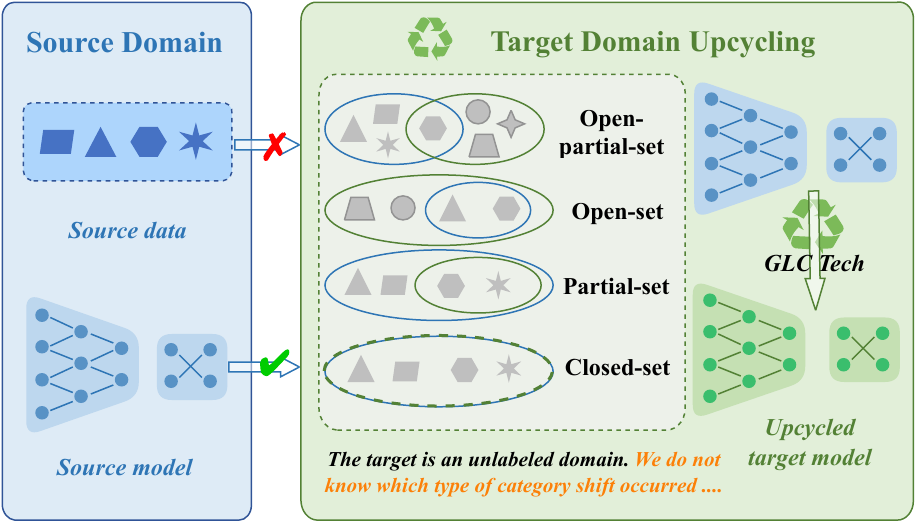}
    \vspace{-0.10in}
    \caption{The illustration of Source-Free Universal Domain Adaptation (SF-UniDA). The goal is to realize model upcycling under both covariate shift and category shift. It is extremely challenging as only one source closed-set model is provided as supervision rather than raw data. Especially, we do not have any prior knowledge about category shifts between domains in advance, e.g., information on matched common classes or the number of categories in the target domain.}
    \vspace{-0.20in}
    \label{fig:sfunida_illustraion}
\end{figure}

\par To tackle these limitations and address those category shift scenarios in a unified manner, in this paper, we take one step further and delve into the Source-Free Universal Domain Adaptation (SF-UniDA). The primary goal is to upcycle the standard pre-trained source models for the target domain under both covariate and category shifts in the absence of direct access to source raw data. We conceptually present the SF-UniDA in Fig.~\ref{fig:sfunida_illustraion}. Note that, very few works~\cite{kundu2020universal, liang2021umad} have studied the source-free domain adaptation (SFDA) in some specific category shift scenarios. Nevertheless, their requirement for dedicated model architectures and prior knowledge about category shifts, greatly limits their practical applications. SF-UniDA is appealing in view that model adaptation can be resolved under various category shifts only on the basis of a standard pre-trained closed-set model, i.e., without specified model architectures.

\par In this paper, to approach such a challenging DA setting, we first propose a simple yet generic technique, \emph{Global and Local Clustering (GLC)}. Different from existing pseudo-labeling strategies that focus on closed-set scenarios, we develop a novel one-vs-all global clustering based pseudo-labeling algorithm to achieve "known" data identification and "unknown" data rejection. As we have no prior about the category shift, we utilize the Silhouettes~\cite{silhouettes} metric to help us realize adaptive global clustering. To avoid source private categories misleading, we design a global confidence statistics based suppression strategy. Although the global clustering algorithm encourages the separation of "known" and "unknown" data samples, some semantically incorrect pseudo-label assignments may still occur, leading to negative knowledge transfer. To mitigate this, we further introduce a local k-NN clustering strategy by exploiting the intrinsic consensus structure of the target domain.

\par Despite the promising results achieved using the simple technique described above, GLC still performs poorly in recognizing different classes of target-private "unknown" data due to the limitations of the closed-set model structure, where "unknown" data are treated as a whole. To alleviate this circumstance, we promote GLC to GLC++ by introducing a new contrastive affinity learning strategy to facilitate the differentiation and identification of clusters. It is worth mentioning that, this strategy is not only beneficial to GLC, but also offers significant advantages when integrated with existing methods, e.g., OVANet~\cite{ova} and UMAD~\cite{liang2021umad}.

\par We have analyzed and validated the superiority of our GLC and its enhanced version GLC++, through extensive experiments on four benchmark datasets (Office-31~\cite{office31}, Office-Home~\cite{officehome}, VisDA~\cite{visda}, and Domain-Net~\cite{domainnet}) under a range of category shift situations, including PDA, OSDA, OPDA and the vanilla CLDA. Additionally, we perform experiments on more challenge benchmarks, including cross-domain semantic segmentation and open-set wild-animal classification. Extensive empirical results demonstrate that GLC and GLC++ yield state-of-the-art performance across multiple benchmarks, even under more stringent conditions.
\par The main contributions are summarized as follows:
\begin{itemize}
    \item To the best of our knowledge, GLC is the first to exploit and achieve the Source-Free Universal Domain Adaptation (SF-UniDA) with only a standard pre-trained closed-set model. 
    \item GLC is equipped with an innovative one-vs-all global clustering based pseudo-labeling algorithm to realize "known" and "unknown" data sample separation under various category shifts.
    \item We introduce a new contrastive affinity learning strategy to enhance the ability to recognize different classes among target-private "unknown" data. It is also complementary to most existing methods, e.g., OVANet~\cite{ova} and UMAD~\cite{liang2021umad}.
    \item Extensive experiments on four benchmarks under various category-shift situations demonstrate the superiority of our GLC and GLC++ techniques. Remarkably, in OPDA scenarios, GLC and GLC++ attain an H-score of 73.2\%/75.3\% on the VisDA benchmark, which are 16.8\%/18.9\% higher than GATE~\cite{gate} respectively.
\end{itemize}

\par This paper is an extended version of our earlier conference submission~\cite{qu2023_glc}. We have extended both the technical contribution and experiment analysis. We here summarize the main extensions: (a) we conduct a more comprehensive literature review; (b) we introduce a new contrastive learning strategy to enhance the capability of novel categories discovery among target "unknown" data; (c) we expand the experimental evaluation by adding a new metric about novel category discovery; (d) we also incorporate our new contrastive learning strategy with existing methods (OVANet~\cite{ova} and UMAD~\cite{liang2021umad}), experiments demonstrate its effectiveness and versatility; (e) we provide a theoretical analysis of the effectiveness and robustness of our methods; (f) we conduct additional experiments on more challenging tasks, such as cross-domain semantic segmentation and open-set wild-animal classification; (g) we provide a more detailed analysis to evaluate GLC/GLC++, including training stability, hyper-parameter sensitivity, and qualitative results.

\section{Related Work}
\subsection{Unsupervised Domain Adaptation}
\par In the field of transfer learning~\cite{pan2009_survey_TL}, unsupervised domain adaptation (UDA) stands out as a typical example, drawing extensive interest in recent years. Its primary goal is to address performance degeneration resulting from covariate shifts. This objective is achieved by harnessing the knowledge contained in a labeled but different source dataset to facilitate the learning of a discriminative model for the unlabeled target dataset. Adhering to the principle of aligning features between source and target domains, existing methods can be broadly classified into three distinct categories: discrepancy-based, reconstruction-based, and adversarial-based. Discrepancy-based approaches~\cite{MMD, zellinger2017_CMD, wasserstein} usually introduce a divergence criterion to measure the distance between source and target data distributions, and then achieve model adaptation by minimizing the corresponding criterion. Several preferred metrics for this purpose include the Maximum Mean Discrepancy (MMD)~\cite{MMD}, the High-Order Central Moment Discrepancy (CMD)~\cite{zellinger2017_CMD}, the Contrastive Domain Discrepancy (CDD)~\cite{constrast_da}, and the Wasserstein metric~\cite{wasserstein}. Reconstruction-based methods~\cite{recon_da_1, recon_da_2, recon_da_3} typically introduce an auxiliary image reconstruction task that guides the network to extract domain-invariant features for model adaptation. Inspired by generative adversarial nets (GAN)~\cite{goodfellow2020_gan}, adversarial-based approaches~\cite{dann, cdan, MCD} leverage domain discriminators to learn domain-invariant features. Despite their proven effectiveness in diverse applications, such as object recognition~\cite{MCD, constrast_da, chen2022_reusing}, semantic segmentation~\cite{dacs_da_seg, hoyer2022_hrda_seg, chen2022deliberated_seg}, and object detection~\cite{liu2022_da_detection, da_object_1, da_object_2}, the majority of existing methods focus only on the vanilla closed-set scenarios, i.e., assuming the label spaces are identical across the source and target domains, which is unpractical and significantly limiting the applicability.

\subsection{Universal Domain Adaptation} 
\par To address the challenge of inconsistent label spaces between source and target domains, various studies have introduced methods tailored to specific scenarios, such as partial-set domain adaptation (PDA)~\cite{pda, cao2018partial, ba3us}, open-set domain adaptation (OSDA)~\cite{panareda2017open, osbp, fang2020_openset, kundu2020_osda, PSDC}, and open-partial-set domain adaptation (OPDA)~\cite{uan, ova, yang2022_onering, liang2021umad}. While these approaches constitute notable progress in addressing discrepancies in label spaces, their applicability is often confined to the scenarios for which they were specifically designed. For example, as a method designed for OPDA, \cite{uan} even underperforms the source model in the partial-set scenario. Since the target domain is unlabeled, it is impractical to ascertain any prior knowledge about category shifts beforehand in real-world applications. In response to this issue, the concept of universal domain adaptation (UniDA)~\cite{dance, dcc, gate} has been introduced. UniDA aims to provide a comprehensive solution for domain adaptation, accommodating all possible label shift scenarios, thereby offering a more versatile and practical solution. Despite the progress in UniDA, most existing methods need access to source data during domain adaptation, which is inefficient and may violate the increasing data protection policies~\cite{GDPR_EU}.

\subsection{Source-Free Domain Adaptation}

\par Source-Free Domain Adaptation (SFDA)~\cite{shot, gsfda, BMD, liang2021_shot_plus, qu2024_lead, qu2025_general_bmd} focuses on achieving domain adaptation by leveraging knowledge from pre-trained source models, in the absence of access to raw source data during adaptation. Existing SFDA methodologies can be categorized into two main types: data generation methods and model fine-tuning methods. Data generation methods~\cite{3c-gan, ye2021_sfda_seg, liu2021_sfda_seg} aim to synthesize source-like image data, facilitating cross-domain adaptation by applying conventional UDA techniques. Model fine-tuning methods~\cite{shot, gsfda, liang2021_shot_plus, nrc} seek to adapt a pre-trained source model to the target domain by utilizing unlabeled target data within a self-supervised training framework. Specifically, model fine-tuning approaches directly fine-tune the models' parameters to improve their performance on the target domain, by leveraging the intrinsic characteristics of the target data for adaptation. However, to avoid model collapse, existing SFDA methods commonly focus on the vanilla closed-set domain adaptation, significantly limiting their usability. Recently, few works~\cite{kundu2020universal, liang2021umad} have studied the source-free domain adaptation in OPDA scenarios. Nevertheless, the requirement of dedicated source model architectures, e.g., specialized two-branch open-set recognition frameworks, greatly limits their deployment in reality. In this paper, we target to achieve source-free universal domain adaptation (SF-UniDA), including PDA, OSDA, OPDA, and the CLDA, with the knowledge from a vanilla source closed-set model.

\subsection{Contrastive Learning}
\par Contrastive learning~\cite{oord2018_contrast_1, tian2020_contrast_2, he2020_contrast_moco, hjelm2018_contrast_3, chen2020_contrast_simclr} has emerged as a prominent approach for its capacity to learn representations without human annotations, marking a significant stride for self-supervised/unsupervised representation learning. It works with a certain dictionary look-up mechanism, in which an image undergoes augmentation to generate two views, query and key. The objective is to enable the query to identify its associated positive key from a set of negative keys, which are derived from disparate images. This technique has been recently integrated into several SFDA methods~\cite{zhang2022_con_sfda_1, liu2022_con_sfda_2, huang2021_con_sfda_3, qiu2021_con_sfda_4}. Despite progress, these approaches predominantly focus on learning correspondence based on the categories of objects present in different views of an image, which restricts the applicability of the learned representations under domain shift. Since they do not adequately capture semantic concepts across the entire dataset, it further becomes pronounced in the context of discovering novel categories within "unknown" data. In this paper, we draw inspiration from contrastive learning and construct contrastive pairs according to their distance in the manifold space, aiming to improve the differentiation and identification of clusters.

\begin{figure*}[ht]
    \centering
    \includegraphics[width=0.99\textwidth]{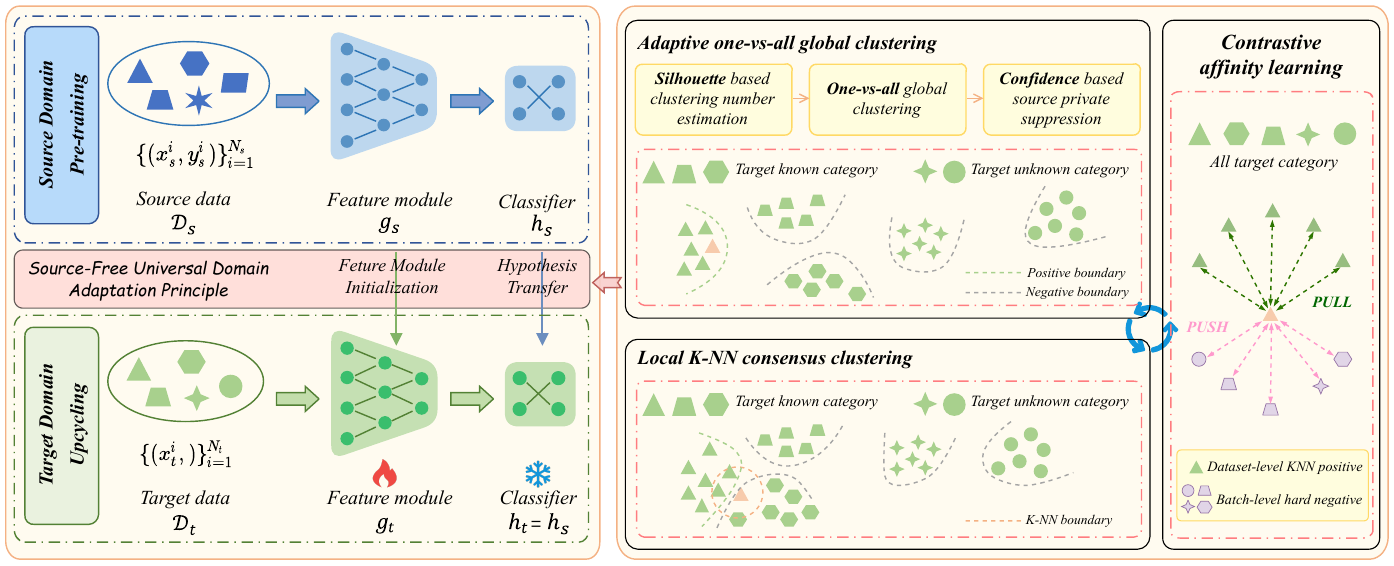}
    \vspace{-0.15in}
    \caption{ Overview of our technique for Source-Free Universal Domain Adaptation (SF-UniDA). In SF-UniDA, the source data $\mathcal{D}_{s}$ serves solely for pre-training. Target model adaptation/upcycling is achieved through the exclusive use of the unlabeled target data $\mathcal{D}_{t}$ and the pre-trained source model. The main challenge is to recognize target data belonging to source-target shared "known" categories, and separate them from target-private "unknown" data. Deviating from existing methods that rely on customized source model architectures, we propose to address SF-UniDA on the basis of a standard closed-set model. Following existing source-free closed-set domain adaptation (SFDA) methods \mbox{\cite{shot, liang2021_shot_plus, BMD}}, given a pre-trained source model $f_s = h_s \circ g_s$, we freeze the classifier $h_s$ and merely learn the target-specific feature module $g_t$ by fine-tuning the source feature module $g_s$ for domain alignment. To realize "known" and "unknown" data separation, we first develop a novel adaptive one-vs-all global clustering algorithm to assign pseudo labels for each target data sample. As we have no prior about the category shift, we introduce the Silhouette \mbox{\cite{silhouettes}} criterion to facilitate us in achieving adaptive one-vs-all clustering. To avoid misleading from source private categories, we develop a global confidence score based suppression strategy. In addition to the global clustering, we also exploit the local intrinsic structure to mitigate negative transfer. The combination of global and local clustering strategies is designated as GLC. Due to the closed-set source design, GLC considers all private target data as a whole, diminishing its effectiveness in identifying distinct categories among the "unknown" data. To elevate its capacity for novel category discovery, we further promote GLC to GLC++, by integrating a new contrastive affinity learning strategy. Note that these three strategies collectively address key challenges in SF-UniDA, each playing a distinct but complementary role in improving the robustness and performance of our approach. In the illustration: blue samples represent source-domain data with ground-truth labels (used only during source pre-training); green samples denote unlabeled target-domain data containing both known and unknown classes; and pink samples are batch-level hard negatives—examples close to an anchor in the mini-batch that do not belong to the same class; and the orange sample, selected from the green target data, is used to demonstrate the role of the three proposed technical components.
    }
    \label{fig:sfunida_framework}
    \vspace{-0.20in}
\end{figure*}

\subsection{Novel Category Discovery}
\par Novel Category Discovery (NCD) proposed in DTC~\cite{han2019_dtc} seeks to categorize unlabeled datasets by utilizing knowledge derived from labeled datasets, whereby the unlabeled dataset consists of classes disjoint to those present in the labeled dataset. Traditional approaches, such as KCL~\cite{hsu2018_kcl} and MCL~\cite{hsu2019_mcl}, typically employ dual models trained separately on labeled and unlabeled data to facilitate general task transfer learning. \cite{zhao2021_ncd_dual_branch} introduces a dual-branch model architecture, comprising two branches dedicated to the learning of global and local features, respectively. This architecture enables dual ranking statistics and mutual learning across the branches, thereby enhancing the efficacy of representation learning and facilitating the discovery of novel categories. Recently, NCD has been expanded into Generalized Category Discovery (GCD)~\cite{vaze2022_gcd}, wherein the unlabeled dataset encompasses a mix of both previously "known" (labeled) and "unknown" (unlabeled) categories. Despite the conceptual overlap with UniDA/SF-UniDA setting, a notable limitation of existing NCD/GCD approaches is their inherent assumption of a non-existent domain gap between the labeled and unlabeled datasets. Moreover, these approaches typically necessitate simultaneous access to both datasets during the learning phase. In this paper, we develop a versatile framework for SF-UniDA, capable of recognizing "known" data, excluding "unknown" data, and uncovering novel categories, despite the presence of domain and category shifts.

\section{Methodology}
\subsection{Preliminary}
\par In this paper, we aim to address a generalized and challenging case in source-free domain adaptation (SFDA): source-free universal domain adaptation (SF-UniDA). The goal is to upcycle models under both covariate and category shifts. In particular, we consider the fundamental yet representative classification task. Formally, we are given a well-designed source domain $\mathcal{D}_s = \{(x^i_s, y^i_s)\}^{N_s}_{i=1}$ where $x^i_s \in \mathcal{X}_s$, $y_s^i \in \mathcal{Y}_s$, and an unlabeled target domain $\mathcal{D}_t = \{(x_t^i, ?)\}^{N_t}_{i=1}$ where $x_t^i \in \mathcal{X}_t$. For a better illustration, we denote $\mathcal{Y} = \mathcal{Y}_s \cap \mathcal{Y}_t$ as the common label space, $\bar{\mathcal{Y}}_s = \mathcal{Y}_s \setminus \mathcal{Y}$ as the source private label space, and $\bar{\mathcal{Y}}_t = \mathcal{Y}_t \setminus \mathcal{Y}$ as the target private label space, respectively. Generally, there are four possible scenarios in SF-UniDA, i.e., the partial-set domain adaptation (PDA), where $\mathcal{Y}_s \supset \mathcal{Y}_t$; the open-set domain adaptation (OSDA), where $\mathcal{Y}_s \subset \mathcal{Y}_t$; the open-partial-set domain adaptation (OPDA), where $\mathcal{Y} \ne \emptyset$, $\bar{\mathcal{Y}}_s \ne \emptyset$, $\bar{\mathcal{Y}}_t \ne \emptyset$, and the well-studied closed-set domain adaptation (CLDA), where $\mathcal{Y}_s = \mathcal{Y}_t$. Given the target domain is unlabeled, SF-UniDA assumes that there is no prior knowledge regarding $\mathcal{Y}_t$. Consequently, both the shared label space $\mathcal{Y}$ and the exclusive label space of the target domain $\bar{\mathcal{Y}}_t$ remain unknown.

\par There have been few works~\cite{kundu2020universal, liang2021umad} explored SFDA under category shift. However, these methods are limited to specific category shift, and require dedicated source model architectures. To address these limitations, we propose to achieve SF-UniDA
through a vanilla closed-set model, diverging from the need for specialized architectures. Following existing closed-set SFDA methods~\cite{shot, BMD, liang2021_shot_plus}, the source model $f_s$ parameterized by a deep neural network consists of two modules: the feature encoding module $g_s$ and the classifier module $h_s$, i.e., $f_s = h_s \circ g_s$. We prepare $f_s$ with the following training objective:
\begin{align}
    \mathcal{L}_{src} = -\frac{1}{N}\sum_{i=1}^{N}\sum_{c=1}^{C_s} q_c \log \delta_c(f_s(x^i_s))
\end{align}
where $\delta_c (f_s(x_s))$ denotes the softmax probability of source sample $x_s$ belonging to the $c$-th category, $q_c$ is the smoothed one-hot encoding of $y_s$, i.e., $q_c = (1 -\alpha)*\mathds{1}_{[c=y_s]} + \alpha / C_s$, and $\alpha$ is a smoothing parameter with a default value of 0.1.

\par To realize model adaptation, we capitalize on the source hypothesis~\cite{shot, liang2021_shot_plus} to achieve source and target domain alignment. That is, we only learn a target-specific feature module $g_t$ and keep the classifier $h_t = h_s$. To realize "known" data identification and "unknown" data rejection under both covariate shift and category shift, we devise a novel, adaptive, one-vs-all global clustering algorithm. Besides, we also employ a local k-NN clustering strategy to alleviate negative transfer. We refer to the combination of our global and local clustering strategies as GLC. We further promote GLC to GLC++ by incorporating a new contrastive affinity learning strategy. The overall pipeline is presented in Fig.~\ref{fig:sfunida_framework}. More details will be described in the following.

\subsection{One-vs-all Global Clustering}
\par Pseudo-labeling is a promising technique in unsupervised domain adaptation. Traditional pseudo-labeling strategies~\cite{psd_label} assign pseudo labels directly based on sample-level predictions, which are often noisy, especially in the presence of covariate shifts. To mitigate this, there have been some pseudo-labeling strategies~\cite{shot, BMD, tpn,  zhang2021_proda} exploit the data structure of the target domain, \emph{i.e.}, the target-specific prototypes. However, these strategies typically assume that source and target domains share identical label space, making it infeasible under category shift.  Consequently, an essential question emerges: \emph{How can we achieve effective pseudo-labeling across inconsistent label spaces, particularly in SF-UniDA where the prior about category shift between $\mathcal{Y}_s$ and $\mathcal{Y}_t$ is inaccessible?}

\par {To tackle this, we first view this problem from a simplified perspective: \emph{If $\mathcal{Y}_s \subset \mathcal{Y}_t$ (i.e., the OSDA setting), and we were to know the number of categories in the target domain is $C_t$, what kind of pseudo-labeling strategy should we apply?}} Intuitively, target domain, in this case, should be grouped into $C_t$ clusters, each corresponding to a specific category. We can then assign pseudo labels via the nearest cluster centroid classifier. However, even though we apply existing clustering algorithms, such as K-means~\cite{kmeans}, to divide the target domain into $C_t$ clusters. It is still challenging to associate the corresponding semantic category for each cluster, as we have no access to the labeled source data.

\par To ease the challenging semantic association, we devise a novel one-vs-all global clustering pseudo-labeling algorithm. The main idea is that \emph{For a particular "known" category $c \in C_s$, in order to decide whether a data sample belongs to the $c$-th category, we need to figure out what is and what is not the $c$-th category.} The procedure is detailed as follows:
\begin{itemize}
    \item For a particular $c$-th category, we first identify and aggregate instances from the target domain $\mathcal{D}_t$ with the highest top-$K$ $\delta_c({f}_t(x_t))$ scores to construct the positive subset $\mathcal{P}_c$, while designating the remainder as the negative subset $\mathcal{N}_c$. Here, $\delta_c({f}_t(x_t))$ denotes the soft-max probability of target instance $x_t$ belonging to the $c$-th class. We empirically set $K = N_t / C_t$.
    \item Then, we obtain the positive prototype ${p}_c$ (\emph{i.e.}, what is the $c$-th category), and negative prototypes $\{{n}_c^{i}\}^{M}_{i=1}$ (\emph{i.e.}, what are not the $c$-th category) via K-means. Noting that we have employed multiple prototypes to represent the negatives since the negatives contain distinct classes.
    We assign $M$ to the value of $C_t$, rather than $C_t - 1$, recognizing that "known" categories in the target domain typically include hard samples, which are challenging to identify with top-$K$ sampling.
    \begin{equation}
        \begin{aligned}
        {p}_c &= \frac{1}{K}\sum_{x_t\in\mathcal{P}_c} g_t(x_t),
        \\
     \{{n}_c^{i}\}_{i=1}^{M} &= \mathop{Kmeans}_{x_t \in \mathcal{N}_c} ({g}_t(x_t)).\\
        \end{aligned}
    \end{equation}

    \item Thereafter, we can decide whether data sample $x_t$ belongs to the $c$-th category via the nearest centroid classifier. Formally,
    \begin{equation}
    \hat{q}_c = \left\{
    \begin{aligned}
     1,\text{if $ S(g_t(x_t), {p}_c) \ge \max \{S(g_t(x_t), {n}_c^i)\}_{i=1}^M$}\\
     0,\text{if $ S(g_t(x_t), {p}_c) < \max \{S(g_t(x_t), {n}_c^i)\}_{i=1}^{M}$}
    \end{aligned}
    \right.
    \label{equ:psd_0}
    \end{equation}
    where $\hat{q}_c = 1$ represents that data sample $x_t$ is pseudo-labeled to the $c$-th category. $S(, )$ denotes the cosine similarity function. 
    \item Finally, we iterate the above process to obtain the pseudo labels $\hat{y}_t$ for all "known" categories $c \in C_s$. Since each data sample either belongs to the "unknown" or one of the categories in the source domain, it is not possible to belong to multiple categories at the same time. Thereby, we introduce a filtering strategy to avoid semantic ambiguity. Here, we just set the category with maximum similarity as the target. It is worth noting that our algorithm does not require the above pseudo-label $\hat{y}_t$ to be one-hot encoded. Those pseudo labels with all-zero encoding mean that these data samples belong to the "unknown" target-private categories $\bar{\mathcal{Y}}_t$. To realize "known" and "unknown" separation, we then manually set those all-zero encoding pseudo labels to a uniform encoding, \emph{i.e.}, $\hat{q}_c = 1/C_s$.
\end{itemize}
\par With the derived pseudo labels, we employ the cross-entropy loss for unsupervised model adaptation. Specifically, 
\begin{equation}
    \begin{aligned}
       \mathcal{L}_{tar}^{glb} &= -\frac{1}{N}\sum_{i=1}^{N}{\sum_{c=1}^{C_s} \hat{q}_c^i}\log \delta_c(h_t(g_t(x_t^i))))
    \end{aligned}
\end{equation}
where $\hat{q}_c^i$ denotes the pseudo label for data sample $x_t^i$.

\subsection{Confidence based Source-private Suppression}

\par In the above section, we developed the one-vs-all global clustering algorithm to assign pseudo labels in OSDA scenarios, when the number of categories in the target domain $C_t$ is available. However, in addition to OSDA, we may also encounter PDA and OPDA, where the source domain contains categories absent in the target domain. To make the above algorithm applicable to OSDA, PDA, and OPDA, it is necessary to tailor it to prevent those source-private categories from misleading pseudo-label assignments.

\par We empirically found that in the positive data group $\mathcal{P}$ sampled with top-K strategy, source-private categories generally yield reduced mean prediction confidence compared to source-target shared categories. Stemming from this observation, we design a suppression strategy for source-private categories based on the mean prediction confidence of the positive data group $\mathcal{P}$. Specifically, for a particular category $c\in C_s$, we tailored the Eq.~\ref{equ:psd_0} to:
\begin{align}
 \epsilon_c &= \rho + \frac{1-\rho}{K}\sum_{x_t\in\mathcal{P}_c} \delta_c(f_t(x_t)), \nonumber\\
\hat{q}_c &= \left\{
\begin{aligned}
 1,\text{if $\epsilon_c\cdot S(g_t(x_t), p_c) \ge \max \{S(g_t(x_t), n_c^i)\}_{i=1}^M$}\\
 0,\text{if $\epsilon_c\cdot S(g_t(x_t), p_c) < \max \{S(g_t(x_t), n_c^i)\}_{i=1}^{M}$}
\end{aligned}
\right.
\label{equ:source_private_suppression}
\end{align}
where $\epsilon_c$ is the designed source-private suppression weight for the $c$-th category, and $\rho$ is a hyper-parameter to control this weight. We empirically set $\rho$ to 0.75 for all benchmarks. Its sensitivity analysis can be found in the experiment.

\subsection{Silhouette-based Target Domain $C_t$ Estimation}

\par Drawing upon the discussions in preceding sections, we have successfully developed the pseudo-labeling algorithm for SF-UniDA. However, its application remains constrained due to the necessity of prior knowledge, i.e., the count of categories $C_t$ within the target domain, which is often elusive in practical scenarios. Thus, the last obstacle for us is: \emph{How to determine the quantity of categories $C_t$ in the target domain?}
\par To address this, a feasible solution is to first enumerate the possible values of the number of categories $C_t$ in the target domain and divide the target domain into the corresponding clusters by applying a clustering algorithm like K-means~\cite{kmeans}. Then the clustering evaluation criteria~\cite{silhouettes, Calinski_Harabasz, Davies_Bouldin, gap_statistic} can be employed to determine the appropriate number of target domain categories $\tilde{C}_t$.
\par In this paper, we simply employ the Silhouette criterion~\cite{silhouettes} to facilitate estimating $\tilde{C}_t$. Technically, for a data sample $x_t \in \mathcal{C}_I$, the Silhouette value $s(x_t)$ is defined as:
\begin{equation}
    \begin{aligned}
     a(x_t) &= \frac{1}{|\mathcal{C}_I| - 1}\sum_{x \in \mathcal{C}_I,  x \ne x_t} d(x_t, x),\\
     b(x_t) &= \min_{J\neq I}\frac{1}{|\mathcal{C}_J|}\sum_{x \in \mathcal{C}_J} d(x_t, x),\\
     s(x_t) &= \frac{b(x_t) - a(x_t)}{\max\{a(x_t), b(x_t)\}}.
\end{aligned}
\end{equation}
where $a(x_t)$ and $b(x_t)$ measure the similarity of $x_t$ to its own cluster $\mathcal{C}_I$ (cohesion) and other clusters $\mathcal{C}_{J, J\neq I}$ (separation), respectively. $d(x_i, x_j)$ measures the distance between data points $x_i$ and $x_j$, and $|\mathcal{C}_I|$ denotes the size of cluster $\mathcal{C}_I$. The Silhouette value $s(x_t)$ ranges from -1 to +1, where a high value indicates that the data sample $x_t$ has a high match with its own cluster and a low match with neighboring clusters. Therefore, if most of the data samples have high Silhouette values, then the clustering configuration is appropriate; otherwise, the clustering configuration may have too many or too few clusters. 
\par Given the complexity and computational demands of accurately determining the number of target categories $C_t$, in our implementation, we have empirically enumerated potential values for $\tilde{C}_t$ as $[1/3C_s, 1/2C_s, C_s, 2C_s, 3C_s]$, taking into account the scenarios may encounter. For efficiency, the value of $\tilde{C}_t$ is estimated initially and maintained constant thereafter. It is anticipated that a more precise estimation $\tilde{C}_t$ would likely improve the accuracy of our one-vs-all global clustering, facilitating better distinction between "known" and "unknown" categories, such refinement is beyond the scope of this work.

\begin{figure}[t]
    \centering
    \includegraphics[width=0.48\textwidth]{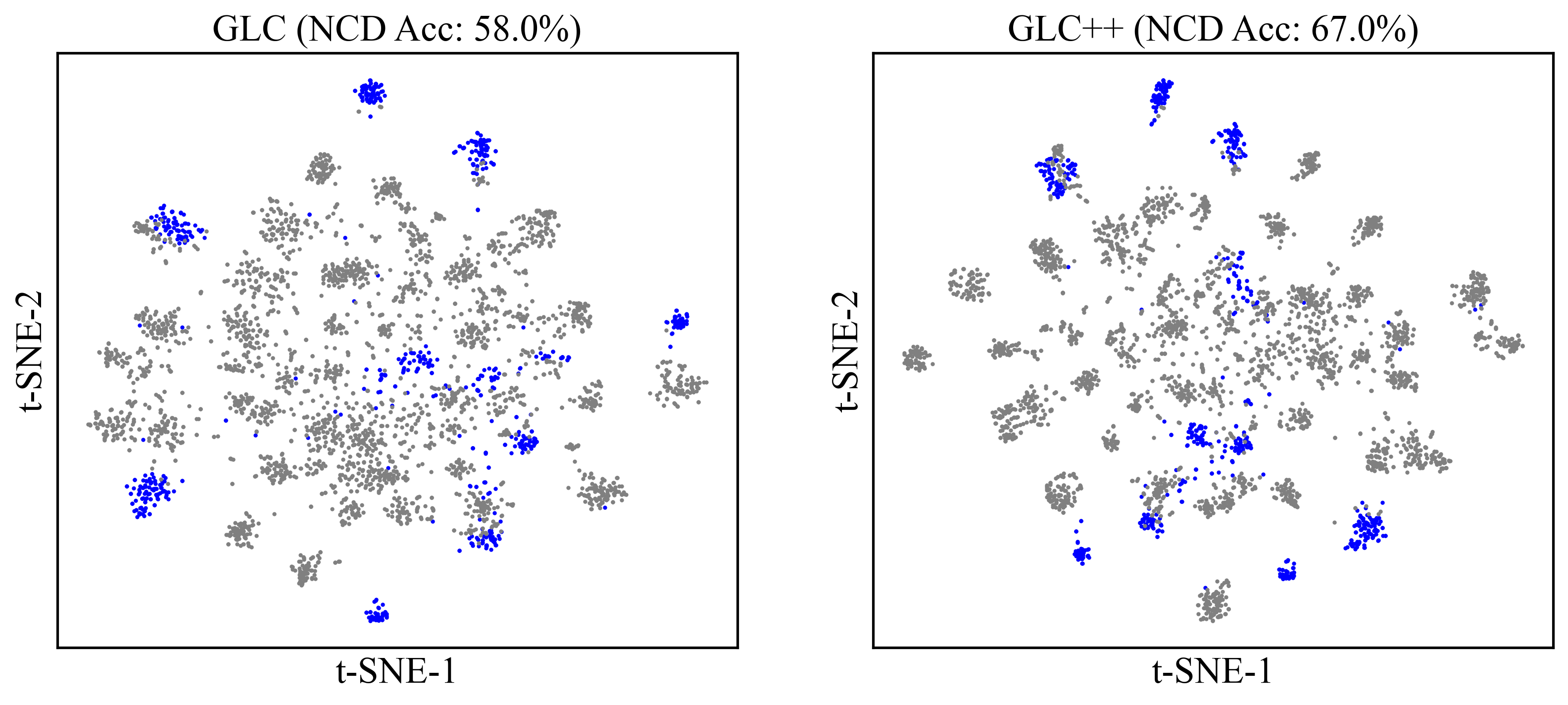}
    \vspace{-0.10in}
    \caption{The t-SNE feature visualization for target-common "known" (blue points) and target-private "unknown" data (gray points) using GLC and GLC++ in the OPDA task Ar$\rightarrow$Pr on the Office-Home dataset. An observation is that GLC exhibits limitations in distinguishing different categories among "unknown" data. This limitation may stem from the inherent constraints of closed-set source architecture, where the pseudo-labeling strategy uniformly categorizes all "unknown" data, treating it as a single entity. Nevertheless, the enhanced GLC++ with our contrastive affinity learning significantly mitigates this issue, demonstrating improved separation of different clusters among "unknown" data.}
    \vspace{-0.15in}
    \label{fig:unknown_tsne}
\end{figure}

\subsection{Local Consensus Clustering}
\par Although the one-vs-all global clustering pseudo-labeling algorithm encourages the separation between "known" and "unknown" data samples, semantically incorrect pseudo-label assignments still occur due to covariate shift, category shift, or inaccurate $C_t$ estimation, resulting in negative transfer.

\par To mitigate this, we further introduce a local k-NN consensus clustering strategy that exploits the intrinsic consensus structure of the target domain $\mathcal{D}_t$ to promote prediction consistency among semantically similar samples. Specifically, during model adaptation, we maintain a memory bank $\mathcal{G}_t = \{g_t(x_t), \delta(f_t(x_t))\}_{x_t \in \mathcal{D}_t}$, which contains the target features and corresponding prediction scores. The local k-NN consensus clustering is then realized by:
\begin{equation}
  \begin{aligned}
     l_c^i &= \frac{1}{|{{L}^i}|}\sum_{x_t\in {L}^i} \delta_c(f_t(x_t)),\\
    \mathcal{L}_{tar}^{loc} &= -\frac{1}{N}\sum_{i=1}^{N}\sum_{c=1}^{C_s} l_c^i\log \delta_c(f_t(x_t^i)).
\end{aligned}  
\end{equation}
where $\delta_c(f_t(x_t))$ denotes the soft-max probability of data instance $x_t$ belonging to the $c$-th class, $L^i$ refers to the set of nearest neighbors of data $x_t^i$ in the embedding feature space. Here, we apply the cosine similarity function to find the nearest neighbors ${L}^i$ of $x_t^i$ in the memory bank $\mathcal{G}_t$. We then encourage minimizing the cross entropy loss between $x_t^i$ and the nearest neighbors ${L}^i$ to achieve the local semantic consensus clustering.

\subsection{Contrastive Affinity Learning }

\par Closed-set source models, advantageous for their architectural flexibility in SF-UniDA, confront an intrinsic challenge with the uniform treatment of target-private "unknown" data, which not only diminishes their ability to distinguish between "known" and "unknown" data but also impedes the categorization of "unknown" data into distinct clusters. Despite the global and local clustering (GLC) technique achieving commendable performance in separating "known" and "unknown" data, its reliance on pseudo-labeling supervision, especially using uniform encoding for all "unknown" data limits its capacity to discriminate among different "unknown" categories. {This limitation is subtly suggested by Fig. \mbox{\ref{fig:unknown_tsne}}, where the adapted model through GLC exhibits more ambiguous boundaries among "unknown" data.}

\par To alleviate this, we promote GLC to GLC++ by developing a new contrastive affinity learning strategy, sidestepping the need for a specialized source model structure. Existing contrastive learning methodologies typically build positive pairs through data augmentation, and designating all other data as negative by default. Despite effectiveness in large-scale unsupervised pre-training, this vanilla strategy restricts its applicability in domain adaptation. It degenerates the semantic concepts across the entire dataset, particularly among the unlabeled "unknown" data. Diverging from conventional approaches, we determine positive and negative pairs based on their proximity within the manifold space. Specifically, we construct positive pairs at the dataset level, utilizing a memory bank established through local consensus clustering, rather than relying on instance-level data augmentation. This design avoids the pitfalls of instance-level data augmentation and maintains the intrinsic semantic structure across the dataset, especially for handling "unknown" data. As for negative pairs, we build a simple yet effective hard-negative mining strategy, instead of treating all the other data in a mini-batch as negative. The rationale behind this is that \emph{hard negative data are a collection of samples that are similar in the embedding space but belong to different categories or clusters.} To mine these samples, we first estimate the expected number of samples for each category in a mini-batch, and then we select the most similar non-expected samples as hard negatives. Given the positive and negative pairs, the contrastive affinity learning is defined as follows:
\begin{align}
&{D}(x_t^i, \mathcal{X}^{i}_{neg}) = \sum_{x \in \mathcal{X}^{i}_{neg}} S(g_t(x^t_i), \texttt{stopgrad}(g_t(x))),\nonumber \\
&{D}(x_t^i, \mathcal{X}^{i}_{pos}) = \sum_{x \in \mathcal{X}^{i}_{pos}} S(g_t(x^t_i), \texttt{stopgrad} (g_t(x))),\nonumber\\
&\mathcal{L}^{con}_{tar} = \frac{1}{N}\sum_{i=1}^{N} {D}(x_t^i, \mathcal{X}^{i}_{neg}) - {D}(x_t^i, \mathcal{X}^{i}_{pos}).
\end{align}
where $\mathcal{X}^{i}_{pos}$ and $\mathcal{X}^{i}_{neg}$ are the positive and negative pairs of data $x_t^i$. $S(,)$ denotes the cosine similarity function. For simplicity, we equalize the size of positive pairs, negative pairs, and local clustering nearest neighbors, i.e., $|\mathcal{X}^{i}_{pos}| = |\mathcal{X}^{i}_{neg}| = |L^i|$. 
In our implementation, we set it to 4 for all benchmark datasets by default. Its sensitivity analysis can be found in the experiments.

\begin{figure*}[ht]
    \centering
    \includegraphics[width=0.90\textwidth]{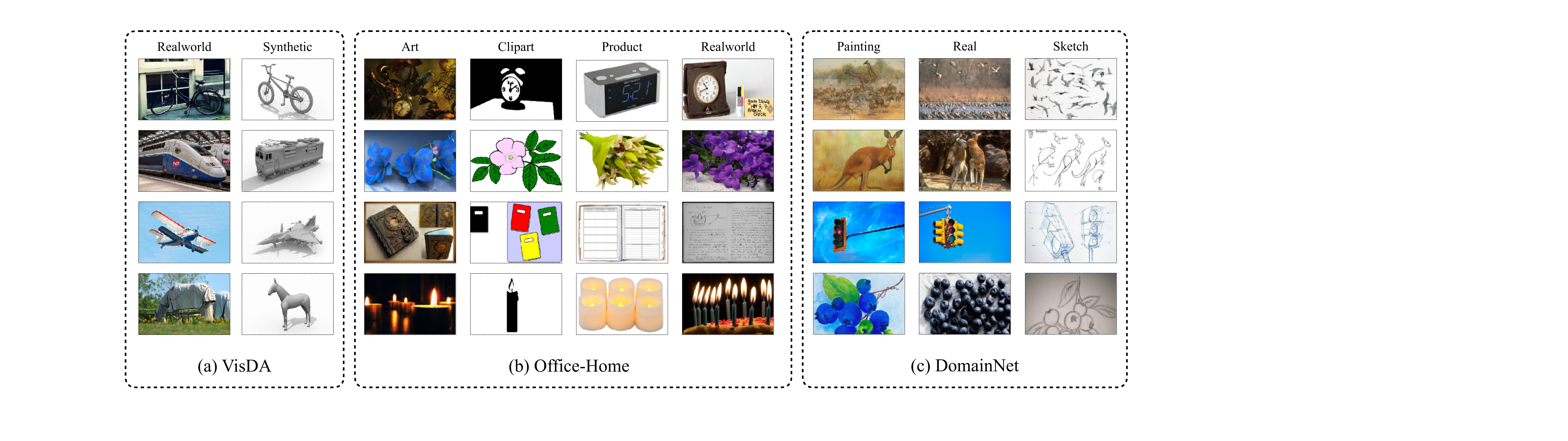}
    \vspace{-0.10in}
    \caption{ Representative examples from the benchmark datasets used in our study, illustrating various types of domain shift. The selected samples highlight the distinct characteristics and environments of the domains within each dataset.
    }
    \vspace{-0.10in}
    \label{fig: dataset}
\end{figure*}

\subsection{Optimization and Inference Details}
\par The overall training loss of GLC and GLC++ can be summarized as:
\begin{align}
   \mathcal{L}_{tar}^{GLC} &= \eta \mathcal{L}_{tar}^{glb} + \mathcal{L}_{tar}^{loc},\\
    \mathcal{L}_{tar}^{GLC++} &= \eta \mathcal{L}_{tar}^{glb} + \mathcal{L}_{tar}^{loc} + \mathcal{L}_{tar}^{con}.
\end{align}
where $\eta > 0$ is a trade-off hyper-parameter.
\par During inference, as there is only one standard classification model, following previous work~\cite{uan, cmu}, we apply the normalized Shannon Entropy~\cite{shannon_entropy} as the uncertainty metric to separate "known" and "unknown" data samples:
\begin{equation}
    I(x_t) = - \frac{1}{\log C_s} \sum_{c=1}^{C_s} \delta_c(f_t(x_t)) \log \delta_c(f_t(x_t))
\end{equation}
where $C_s$ is the class number of source domain $\mathcal{D}_s$, and $\delta_c(f_t(x_t))$ denotes the soft-max probability of data sample $x_t$ belonging to the $c$-th class. The higher the uncertainty, the more the model $f_t$ tends to assign an "unknown" label to the data sample. During inference stage, given an input sample $x_t$, we first compute $I(x_t)$ and then predict the class of $y(x_t)$ with a pre-defined threshold $\omega$ as:
\begin{align}
y(x_t) &= \left\{
\begin{aligned}
 &\text{unknown}, &\text{if $I(x_t) \ge \omega$}\\
 &\text{argmax}(f_t(x_t)), &\text{if $I(x_t) < \omega$}
\end{aligned}
\right.
\end{align}
which either rejects the input sample $x_t$ as "unknown" or classifies it into a "known" class. In our implementation, we set $\omega = 0.55$ for all datasets. 

\subsection{Advancing Existing Methods}
\par We also provide an extension of our contrastive affinity learning strategy for existing baseline methods. Most current techniques developed for OSDA and OPDA ignore the varied semantics present in unlabeled "unknown" data and treat them as a whole, hindering the separation of "known" and "unknown" data, and limiting the discovery of novel categories. Therefore, our contrastive affinity strategy also serves as a beneficial complement to these methods. To validate this merit, we have incorporated our strategy into two representative methodologies, namely source-available OVANet~\cite{ova} and source-free UMAD~\cite{liang2021umad}. For ease of integration, we directly incorporate $\mathcal{L}_{tar}^{con}$ into the baseline optimization objective.
\begin{align}
    \mathcal{L}_{overall} = \mathcal{L}_{baseline} + \gamma\mathcal{L}^{con}_{tar}.
\end{align}
where the selection of $\gamma$ aligns with the baseline methods. Specifically, in the case of OVANet, we set $\gamma = 0.1$, mirroring the baseline's chosen weight for the target-domain adaptation loss. As for UMAD, we adjust $\gamma = 1.0$.

\subsection{Theoretical Analysis}
{Inspired by \mbox{\cite{wei2020_theoretical}}, we leverage the "expansion" and "separation" assumptions to provide a theoretical understanding of our method.}
\\
\noindent {\textbf{Notations}: Let $P$ be the distribution of unlabeled examples over the input space $\mathcal{X}$. We denote by $G^{\star}(x)$ the ground-truth label of $x$, and by $P_i(x)$ the class-conditional distribution of $x$ given $G^{\star}(x) = i$. Let $G_{pl}(x)$ be the pseudo-labeler based on the pre-trained source model, which generates pseudo-labels for unlabeled target data. We define $\mathcal{M}(G_{pl}) \triangleq \{ x : G_{pl}(x) \ne G^\star(x) \}$ as the set of examples that are pseudo-labeled incorrectly. The target classifier we aim to learn from the unlabeled data is denoted by $G$. The disagreement between $G$ and $G^\star$ is given by $\mathrm{Err}(G)\triangleq L_{0-1}(G,G^\star)\triangleq \mathbb{E}_{P}[\mathbf{1}(G(x)\ne G^\star(x))]$. Let $\mathcal{T}$ be a family of data augmentations. We define $\mathcal{B}(x) \triangleq \{ x^{\prime}: \exists\,T\in \mathcal{T},\ \mathrm{such\ that}\ \| x^{\prime} - T(x)\| \le r\}$, and the neighborhood of $x$ as $\mathcal{N}(x) \triangleq \bigl\{ x^{\prime}: \mathcal{B}(x) \cap \mathcal{B}(x^{\prime}) \neq \emptyset \bigr\}$.}
\\
\noindent {\textbf{Assumption 1.1} (Expansion): 
Expansion assumes that data points of the same class form a continuous region and that any sufficiently small region $V$ "expands" to a larger region of the same class. Formally, $(a, z)$-expansion states that for each class-conditional distribution $P_i$, $P_i(\mathcal{N}(V)) \;\ge\; \min \bigl\{ z\,P_i(V), \;1 \bigr\}$, for all $V \subseteq \mathcal{X}$ with $P_i(V) \le a$. Let $\bar{a} \triangleq \max_{i} \bigl\{ P_i(\mathcal{M}(G_{pl})) \bigr\}$ denote the maximum class-wise fraction of incorrectly pseudo-labeled examples. We assume $\bar{a} < {1}/{3}$ and $P$ satisfies $(\bar{a}, \bar{z})$-expansion for $\bar{z} > 3$, the bound for $z$ is $z \;\triangleq\; \min \bigl\{ 1/\bar{a},\, \bar{z}\bigr\}$. Due to the unavailability of target labels, this assumption is not intended to be empirically verifiable. It serves as a theoretical condition that ensures the mislabeling rate among pseudo-labeled examples remains low to maintain the expansion property.}
\\
\noindent {\textbf{Assumption 1.2} (Separation): Separation requires that different class-conditional distributions are well separated by the ground-truth classifier $G^\star$ with high probability. Formally, with probability at least $1 - \epsilon$: $\mathbf{1}(\exists \, x^{\prime} \in \mathcal{B}(x) : G^\star (x^{\prime}) \neq G^\star (x)) \;=\; 0$, which is equivalently stated as $\mathbb{E}_{P}\bigl[\mathbf{1}(\exists\, x^{\prime} \in \mathcal{B}(x) : G^\star (x^{\prime}) \neq G^\star (x))\bigr] \;\le\; \epsilon$. This implies that points from different classes lie at least $2r$ apart (in $l_2$ norm, as used in $\mathcal{B}(\cdot)$). Since $r$ can be much smaller than the typical norm of a sample, the expansion condition here is weaker than a strict clustering assumption, which would require intra-class distances to be strictly smaller than inter-class distances.}
\\
\noindent {\textbf{Theorem 1.3}. 
Under Assumptions 1.1 and 1.2, for any target classifier $\hat{G}$, we have:}
\begin{equation}
  {\mathrm{Err}(\hat{G}) \le \frac{2}{z - 1}\mathrm{Err}(G_{pl}) + \frac{2z}{z - 1}\epsilon.}
\end{equation}
\par {This theorem shows that the performance of the target model $\hat{G}$ is thus bounded by the accuracy of the pseudo-labeler $G_{pl}$ and the separation factor $\epsilon$. In our work, we introduce a novel one-vs-all clustering strategy to produce more reliable pseudo-labels. We then incorporate a local k-NN consensus mechanism and a contrastive affinity learning strategy to better separate different classes. These designs help minimize $\mathrm{Err}(\hat{G})$, leading to robust and accurate results for the target model. Rigorous proof of the Theorem 1.3 can be found in~\mbox{\cite{wei2020_theoretical}}.}

\section{Experiments}
\subsection{Setup}
\noindent \textbf{Datasets:} We utilize the following standard datasets to evaluate the effectiveness and versatility of our methods. \textbf{Office-31}~\cite{office31} is a widely-used small-sized benchmark, consisting of 31 object classes (4,652 images) under an office environment from three domains (DSLR (D), Amazon (A), and Webcam (W)). \textbf{Office-Home}~\cite{officehome} is another popular medium-sized benchmark, consisting of 65 categories (15,500 images) from four domains (Artistic images (Ar), Clip-Art images (Cl), Product images (Pr), and Real-World images (Rw)). \textbf{VisDA}~\cite{visda} is a more challenging benchmark with 12 object classes, where the source domain contains 152,397 synthetic images generated by rendering 3D models and the target domain consists of 55,388 images from Microsoft COCO. \textbf{Domain-Net}~\cite{domainnet}, is the largest domain adaptation benchmark with about 0.6 million images, which contains 345 classes. Similar to previous works~\cite{gate, liang2021umad}, we conduct experiments on three subsets from it (Painting (P), Real (R), and Sketch (S)). Fig.~\ref{fig: dataset} illustrates some representative examples of these datasets. We evaluate GLC and GLC++ in open-partial-set DA (OPDA), open-set DA (OSDA), partial-set DA (PDA), and closed-set DA (CLDA) scenarios. Table~\ref{tab:label_split} details the class split for each scenario. 

\begin{table}[t]
\centering
\caption{Details of class split for each scenario. Here, $\mathcal{Y}$, $\bar{\mathcal{Y}}_s$, and $\bar{\mathcal{Y}}_t$ denotes the source-target-shared class, the source-private class, and the target-private class, respectively.}
\vspace{-0.05in}
\addtolength{\tabcolsep}{-1.0pt}
\resizebox{0.49\textwidth}{!}{
\begin{tabular}{l|cccc}
\toprule
\multirow{2}{*}{Dataset} & \multicolumn{4}{c}{Class Split ($\mathcal{Y}/ \bar{\mathcal{Y}}_s/ \bar{\mathcal{Y}}_t$)}  \\
\cmidrule{2-5} & OPDA  & OSDA  & PDA & CLDA\\ 
\midrule
Office-31~\cite{office31} & 10/10/11 & 10/0/11 & 10/21/0 & 31/0/0 \\
Office-Home~\cite{officehome} & 10/5/50 & 25/0/40 & 25/40/0 & 65/0/0 \\
VisDA~\cite{visda} & 6/3/3 & 6/0/6  & 6/6/0  & -\\
DomainNet~\cite{domainnet} & 150/50/145   & -       & - & -\\ 
\bottomrule
\end{tabular}
}
\vspace{-0.20in}
\label{tab:label_split}
\end{table}

\begin{table*}[htbp]
  \centering
  \caption{{H-score (\%) comparison in OPDA scenarios on the Office-Home dataset. Some results are cited from GATE~\mbox{\cite{gate}}. SF denotes source data-free. We compare GLC and GLC++ with source-available and source-free methods.}}
  \vspace{-0.05in}
  \addtolength{\tabcolsep}{-3.0pt}
  \resizebox{0.99\textwidth}{!}{
    \begin{tabular}{lccccc|cccccccccccca}
    \toprule
     {Methods} & {SF} & {OPDA} & {OSDA} & {PDA} & {CLDA}  & Ar2Cl & Ar2Pr & Ar2Re & Cl2Ar & Cl2Pr & Cl2Re & Pr2Ar & Pr2Cl & Pr2Re & Re2Ar & Re2Cl & Re2Pr & Avg \\
    \midrule
    UAN~\cite{uan}   & \xmark & \cmark & \xmark & \xmark & \xmark  & 51.6  & 51.7  & 54.3  & 61.7  & 57.6  & 61.9  & 50.4  & 47.6  & 61.5  & 62.9  & 52.6  & 65.2  & 56.6  \\
    CMU~\cite{cmu}   & \xmark & \cmark & \xmark & \xmark & \xmark  & 56.0  & 56.9  & 59.2  & 67.0  & 64.3  & 67.8  & 54.7  & 51.1  & 66.4  & 68.2  & 57.9  & 69.7  & 61.6  \\
    ROS   & \xmark & \xmark & \cmark & \xmark & \xmark  & 54.0  & 77.7  & 85.3  & 62.1  & 71.0  & 76.4  & 68.8  & 52.4  & 83.2  & 71.6  & 57.8  & 79.2  & 70.0  \\
    DANCE~\cite{dance} & \xmark & \cmark & \cmark & \cmark & \cmark & 61.0  & 60.4  & 64.9  & 65.7  & 58.8  & 61.8  & 73.1  & 61.2  & 66.6  & 67.7  & 62.4  & 63.7  & 63.9  \\
    DCC~\cite{dcc}  & \xmark & \cmark & \cmark & \cmark & \cmark & 58.0  & 54.1  & 58.0  & {{74.6}}  & 70.6  & 77.5  & 64.3  & {{73.6}}  & 74.9  & {{81.0}}  & {{75.1}}  & 80.4  & 70.2  \\
    GATE~\cite{gate}  & \xmark & \cmark & \cmark & \cmark & \cmark  & {{63.8}}  & {75.9}  & {81.4}  & {{74.0}}  & {{72.1}}  & {{79.8}}  & {{74.7}}  & {{70.3}}  & {{82.7}}  & 79.1  & {{71.5}}  & {{81.7}}  & {{75.6}}  \\
    \midrule
    OVANet~\cite{ova} & \xmark & \cmark & \cmark & \xmark & \xmark  & 61.7 & 77.3 & 77.9 & 69.7 & 69.2 & 75.7 & 70.6 & 58.2 & 79.2 & 76.8 & 63.0 & 79.0 & 71.6  \\
    &  &  &  & & & $\pm$0.5 &$\pm$0.3 &$\pm$0.3 &$\pm$0.4 &$\pm$0.3 &$\pm$0.3 &$\pm$0.4 &$\pm$0.6 &$\pm$0.2 &$\pm$0.4 &$\pm$0.2 &$\pm$0.3 &$\pm$0.3 \\
    OVANet + $\mathcal{L}_{tar}^{con}$ & \xmark & \cmark & \cmark & \xmark & \xmark  & 62.8 & 76.2 & 79.2 & {{72.7}} & 69.8 & 75.6 & 72.8 & 61.0 & 78.5 & 78.2 & 63.7 & 78.2 &\textbf{72.4} \\
    &  &  &  & & & $\pm$0.3 &$\pm$0.5 &$\pm$0.3 &$\pm$0.8 &$\pm$1.1 &$\pm$0.5 &$\pm$0.3 &$\pm$0.4 &$\pm$0.4 &$\pm$0.5 &$\pm$0.3 &$\pm$0.2 &$\pm$0.5 \\
    \midrule
    \midrule
    Source-only & \cmark & -     & -    & -  & -    & 47.3  & 71.6  & 81.9  & 51.5  & 57.2  & 69.4  & 56.0  & 40.3  & 76.6  & 61.4  & 44.2  & 73.5  & 60.9  \\
    SHOT-O~\cite{shot} & \cmark & \xmark & \cmark & \xmark & \xmark & 32.9  & 29.5  & 39.6  & 56.8  & 30.1  & 41.1  & 54.9  & 35.4  & 42.3  & 58.5  & 33.5  & 33.3  & 40.7  \\
    \midrule
    UMAD~\cite{liang2021umad}  & \cmark & \cmark & \cmark & \xmark & \xmark & 58.7	& 77.1	& 85.4	& 70.0	& 67.6	& 76.3	& 69.0	& 56.8	& 83.0	& 79.5	& 62.0	& 80.8	& 72.2 \\
    &  &  &  & & & $\pm$0.2 &$\pm$0.7 &$\pm$0.3 &$\pm$0.5 &$\pm$0.4 &$\pm$0.3 &$\pm$0.1 &$\pm$0.3 &$\pm$0.5 &$\pm$0.4 &$\pm$0.5 &$\pm$0.2 &$\pm$0.4 \\
    UMAD + $\mathcal{L}_{tar}^{con}$ & \cmark & \cmark & \cmark & \xmark & \xmark & 59.9	& 78.0	& 86.2	& 71.2	& 69.1	& 78.0	& 73.2	& 59.7	& 85.7	& 80.2	& 63.8	& 80.6	& \textbf{73.8}\\
    &  &  &  & & & $\pm$0.4 &$\pm$0.4 &$\pm$0.4 &$\pm$0.2 &$\pm$0.7 &$\pm$0.4 &$\pm$0.3 &$\pm$0.6 &$\pm$0.1 &$\pm$0.2 &$\pm$0.5 &$\pm$0.3 &$\pm$0.4 \\
    \midrule
    GLC   & \cmark & \cmark & \cmark & \cmark & \cmark& {{62.4}}      & {{78.6}}      & {{89.6}}     & 70.0      & {{77.0}}      & {{87.3}}      & {{77.4}}      & 56.9      & {{89.5}}      & {{80.6}}      & 53.6      & {{85.3}}      & {{75.7}} \\
    &  &  &  & & &$\pm$0.7 &$\pm$0.2 &$\pm$0.1 &$\pm$1.0 &$\pm$1.2 &$\pm$2.0 &$\pm$0.7 &$\pm$0.6 &$\pm$0.3 &$\pm$0.3 &$\pm$0.4 &$\pm$0.5 &$\pm$0.7 \\
    GLC++ & \cmark & \cmark & \cmark & \cmark & \cmark & {{69.0}} & {{78.8}} & {{90.0}} & 70.1 & {{81.2}} & {{86.1}} & {{78.8}} & 57.9 & {{89.2}} & {{81.3}} & 54.2 & {{87.0}} & \textbf{77.0} \\
    &  &  &  & & & $\pm$0.9 &$\pm$0.5 &$\pm$0.4 &$\pm$0.1 &$\pm$1.2 &$\pm$1.7 &$\pm$0.3 &$\pm$0.5 &$\pm$0.3 &$\pm$0.3 &$\pm$0.4 &$\pm$0.4 &$\pm$0.6 \\
    \bottomrule
    \end{tabular}
    }
  \label{tab:opda_Office-Home}%
  \vspace{-0.10in}
\end{table*}%

\noindent \textbf{Evaluation protocols:} For a fair comparison, we utilize the same evaluation metric as previous works~\cite{gate, dcc}. Specifically, in PDA and CLDA scenarios, we report the classification accuracy over all target samples. In OSDA and OPDA scenarios, considering the trade-off between "known" and "unknown" categories, we report the H-score, i.e., the harmonic mean of the accuracy of "known" and "unknown" samples. Furthermore, we also introduce the novel category discovery accuracy (\textit{NCD Acc}) as a metric to assess the capability to distinguish between distinct categories among the "unknown" data. Formally, it defined as $\frac{1}{N_t^{unk}}\sum_{i=1}^{N_t^{unk}} \mymathbb{1}(y_i^* = H(y_i))$ where $H$ denotes the optimal permutation derived through the Hungarian algorithm~\cite{kuhn1955_hungarian}. $H$ matches the clustering assignment $y_i$ with the ground-truth "unknown" category label $y^*_i$, $N_t^{unk}$ represents the amount of "unknown" data.
\begin{table*}[htbp]
  \centering
  \caption{{H-score (\%) comparison in OPDA scenarios on the Office-31, VisDA, and DomainNet datasets, respectively.}}
  \vspace{-0.05in}
  \addtolength{\tabcolsep}{-2.0pt}
  \resizebox{0.99\textwidth}{!}{
    \begin{tabular}{lccccc|cccccca|a|cccccca}
    \toprule
    \multirow{2}[4]{*}{Methods} & \multirow{2}[4]{*}{SF} & \multirow{2}[4]{*}{OPDA} & \multirow{2}[4]{*}{OSDA} & \multirow{2}[4]{*}{PDA} & \multirow{2}[4]{*}{CLDA} & \multicolumn{7}{c|}{Office-31}              & \multicolumn{1}{c|}{VisDA} & \multicolumn{7}{c}{DomainNet} \\
\cmidrule{7-21} &     &       &       &       &       & A2D   & A2W   & D2A   & D2W   & W2A   & W2D   & Avg   &  \textbf{S2R}     & {P2R} & {P2S} & {R2P} & {R2S} & {S2P} & {S2R} & Avg \\
    \midrule
    UAN~\cite{uan}   & \xmark & \cmark & \xmark & \xmark & \xmark & 59.7  & 58.6  & 60.1  & 70.6  & 60.3  & 71.4  & 63.5  & 34.8  & {41.9} & {39.1} & {43.6} & {38.7} & {38.9} & {43.7} & {41.0} \\
    CMU~\cite{cmu}   & \xmark & \cmark  & \xmark & \xmark & \xmark & 68.1  & 67.3  & 71.4  & 79.3  & 72.2  & 80.4  & 73.1  & 32.9  & {50.8} & {{45.1}} & {52.2} & {45.6} & {44.8} & {51.0} & {48.3} \\
    ROS~\cite{ros}   & \xmark & \xmark & \cmark  & \xmark & \xmark & 29.8  & 26.8  & 86.4  & 86.6  & 83.9  & 96.0  & 68.3  & 30.3  & {20.5} & {36.9} & {30.0} & {19.9} & {28.7} & {23.2} & {26.5} \\
    DANCE~\cite{dance} & \xmark & \cmark & \cmark & \cmark  & \cmark & 78.6  & 71.5  & 79.9  & 91.4  & 72.2  & 87.9  & 80.3  & 42.8  & {21.0}  & {37.0} & {47.3} & {46.7} & {27.7}  & {21.0} & {33.5} \\
    DCC~\cite{dcc}   & \xmark & \cmark & \cmark & \cmark  & \cmark & {{88.5}}  & 78.5  & 70.2  & 79.3  & 75.9  & 88.6  & 80.2  & 43.0  & {56.9} & {43.7} & {50.3} & {43.3} & {44.9} & {56.2} & {49.2} \\
    GATE~\cite{gate}  & \xmark & \cmark & \cmark & \cmark  & \cmark & {{87.7}}  & {{81.6}}  & 84.2  &{{94.8}}  & 83.4  & 94.1  & {{87.6}}  & 56.4  & 57.4 & {{48.7}} & {{52.8}}  & {{47.6}} & {{49.5}} & {56.3} & {{52.1}} \\
    \midrule
    OVANet~\cite{ova} & \xmark & \cmark & \cmark & \xmark  & \xmark & 84.1 & 79.3 & 78.5	& 94.3	& 83.5	& 95.8	& 85.9 & {48.1} & 55.4 & 45.8 & 52.1 & 44.8 & 46.3 & 54.5 & 49.8\\
    &  &  &  & & & $\pm$0.8 &$\pm$0.6 &$\pm$1.3 &$\pm$0.4 &$\pm$0.9 &$\pm$0.2 &$\pm$0.7 &{$\pm$0.8}   &$\pm$0.5    &$\pm$0.7   &$\pm$0.2   &$\pm$0.2   &$\pm$0.4   &$\pm$0.3 &$\pm$0.4\\
    OVANet + $\mathcal{L}_{tar}^{con}$ & \xmark & \cmark & \cmark & \xmark  & \xmark & 84.5	& 81.1	& 84.3	& 94.7	& 85.1	& 94.3	& \textbf{87.3} & \textbf{53.4} & 55.0 & 46.9 & 52.1 & 45.9 & 46.5 & 54.6 & \textbf{50.2} \\
    &  &  &  & & & $\pm$0.9 &$\pm$1.0 &$\pm$1.1 &$\pm$0.3 &$\pm$0.4 &$\pm$0.6 &$\pm$0.7 &$\pm$0.8 &$\pm$0.4 &$\pm$0.8 &$\pm$0.3 &$\pm$0.1 &$\pm$0.3 &$\pm$0.2 &$\pm$0.3\\
    \midrule
    \midrule
    Source-only & \cmark & -     & -     & -  & -    & 70.9  & 63.2  & 39.6  & 77.3  & 52.2  & 86.4  & 64.9  & 25.7  & 57.3      & 38.2      & 47.8      & 38.4      & 32.2      & 48.2      & 43.7 \\
    SHOT-O~\cite{shot} & \cmark & \xmark & \cmark & \xmark  & \xmark & 73.5  & 67.2  & 59.3  & 88.3  & 77.1  & 84.4  & 75.0  & 44.0  & {35.0} & {30.8} & {37.2} & {28.3} & {31.9} & {32.2} & {32.6} \\
    \midrule
    UMAD & \cmark & \cmark & \cmark & \xmark  & \xmark & 86.9	& 88.3	& 90.9	& 95.9	& 90.0	& 98.0	& {91.7}  & {55.1} & 53.4 & 39.5 & 44.1 & 37.6 & 40.5 & 52.3 & 44.6\\
    &  &  &  & & & $\pm$0.3 &$\pm$1.1 &$\pm$0.4 &$\pm$0.1 &$\pm$0.1 &$\pm$0.4 &$\pm$0.4 &{$\pm$0.3} &$\pm$0.6 &$\pm$0.3 &$\pm$0.2 &$\pm$0.4 &$\pm$0.3 &$\pm$0.5 &$\pm$0.4\\
    UMAD + $\mathcal{L}_{tar}^{con}$ & \cmark & \cmark & \cmark & \xmark  & \xmark & 87.6	& 91.4	& 91.6	& 96.8	& 91.4	& 97.9	& \textbf{92.8}  & \textbf{56.8} &54.5 &40.0 &44.8 & 40.7 &41.5 & 53.3 &\textbf{45.8} \\
    &  &  &  & & & $\pm$0.4 &$\pm$0.6 &$\pm$0.1 &$\pm$0.2 &$\pm$0.0 &$\pm$0.4 &$\pm$0.3 &{$\pm$0.2}  &$\pm$0.5 &$\pm$0.4 &$\pm$0.4 &$\pm$0.3 &$\pm$0.2 &$\pm$0.5 &$\pm$0.4\\
    \midrule
    GLC   & \cmark & \cmark & \cmark & \cmark  & \cmark & {81.0}      & {{83.8}}      &  {{89.3}}     & 90.9      & {{88.0}}      & 92.4      & {{87.6}}      & {73.2} & 63.5 & 50.2 & 55.3 & 50.5 & 49.5 & 61.2 & \textbf{55.0} \\
    &  &  &  & & & $\pm$0.2 &$\pm$0.7 &$\pm$0.2 &$\pm$0.4 &$\pm$0.4 &$\pm$0.1 &$\pm$0.3 &$\pm$0.4 &$\pm$0.3 &$\pm$0.6 &$\pm$0.1 &$\pm$0.5 &$\pm$0.3 &$\pm$0.5 &$\pm$0.3\\
    GLC++  & \cmark & \cmark & \cmark & \cmark  & \cmark & 81.7 & 85.3 & 89.7 & 91.5 & 88.1 & 92.6 & \textbf{88.2} & \textbf{75.3} &62.8 & 50.5 & 55.0 & 50.5 &49.0 &60.9 & 54.8 \\
    &  &  &  & & & $\pm$0.3 &$\pm$1.0 &$\pm$0.4 &$\pm$0.2 &$\pm$0.3 &$\pm$0.2 &$\pm$0.4 &$\pm$0.3 &$\pm$0.4 &$\pm$0.3 &$\pm$0.3 &$\pm$0.2 &$\pm$0.4 &$\pm$0.2 &$\pm$0.3\\
    \bottomrule
    \end{tabular}%
  }
  \label{tab:opda_rest}%
  \vspace{-0.1in}
\end{table*}%
\begin{table*}[htbp]
  \centering
  \caption{{H-score (\%) comparison in OSDA scenarios on the Office-Home, Office-31, and VisDA datasets, respectively.}}
   \vspace{-0.05in}
  \addtolength{\tabcolsep}{-4.0pt}
  \resizebox{0.99\textwidth}{!}{
    \begin{tabular}{lccccc|cccccccccccca|a|a}
    \toprule
    \multirow{2}[4]{*}{Methods} & \multirow{2}[4]{*}{SF} & \multirow{2}[4]{*}{OPDA} & \multirow{2}[4]{*}{OSDA} & \multirow{2}[4]{*}{PDA} & \multirow{2}[4]{*}{CLDA} & \multicolumn{13}{c|}{Office-Home}                                                           & \multicolumn{1}{c|}{Office31}  &
    \multicolumn{1}{c}{VisDA} \\
\cmidrule{7-21}      &    &       &       &       &       & Ar2Cl & Ar2Pr & Ar2Re & Cl2Ar & Cl2Pr & Cl2Re & Pr2Ar & Pr2Cl & Pr2Re & Re2Ar & Re2Cl & Re2Pr & Avg   & Avg   & \textbf{S2R} \\
    \midrule
    STAmax~\cite{liu2019_stamax} & \xmark  & \xmark  & \cmark & \xmark & \xmark & 55.8  & 54.0  & 68.3  & 57.4  & 60.4  & 66.8  & 61.9  & 53.2  & 69.5  & 67.1  & 54.5  & 64.5  & 61.1  & 72.5  & 64.1 \\
    OSBP~\cite{osbp}  & \xmark  & \xmark  & \cmark & \xmark & \xmark  & 55.1  & 65.2  & 72.9  & {{64.3}}  & 64.7  & 70.6  & 63.2  & 53.2  & 73.9  & 66.7  & 54.5  & 72.3  & 64.7  & 83.7  & 52.3 \\
    ROS~\cite{ros}   & \xmark  & \xmark  & \cmark & \xmark & \xmark  & 60.1  & 69.3  & 76.5  & 58.9  & 65.2  & 68.6  & 60.6  & 56.3  & 74.4  & {{68.8}}  & 60.4  & {{75.7}}  & 66.2  & 85.9  & 66.5 \\
    UAN~\cite{uan}   & \xmark  & \cmark & \xmark  & \xmark  & \xmark & 34.7  & 22.4  & 9.4   & 38.9  & 22.9  & 21.8  & 47.4  & 39.7  & 30.9  & 34.4  & 35.8  & 22.0  & 30.0  & 55.1  & 51.9 \\
    CMU~\cite{cmu}   & \xmark  & \cmark & \xmark  & \xmark & \xmark  & 55.0  & 57.0  & 59.0  & 59.3  & 58.2  & 60.6  & 59.2  & 51.3  & 61.2  & 61.9  & 53.5  & 55.3  & 57.6  & 65.2  & 54.2 \\
    DANCE~\cite{dance} & \xmark  & \cmark & \cmark & \cmark & \cmark & 6.5   & 9.0   & 9.9   & 20.4  & 10.4  & 9.2   & 28.4  & 12.8  & 12.6  & 14.2  & 7.9   & 13.2  & 12.9  & 79.8  & 67.5 \\
    DCC~\cite{dcc}   & \xmark  & \cmark & \cmark & \cmark & \cmark & 56.1  & 67.5  & 66.7  & 49.6  & 66.5  & 64.0  & 55.8  & 53.0  & 70.5  & 61.6  & 57.2  & 71.9  & 61.7  & 72.7  & 59.6 \\
    GATE~\cite{gate}  & \xmark  & \cmark & \cmark & \cmark & \cmark & {{63.8}}  & 70.5  & 75.8  & {{66.4}}  & 67.9  & 71.7  & {{67.3}}  & {{61.5}}  & {{76.0}}  & {{70.4}}  & {{61.8}}  & 75.1  & {{69.0}}  & 89.5  & {{70.8}} \\
    \midrule
    OVANet~\cite{ova} & \xmark  & \cmark & \cmark & \xmark & \xmark & 59.3	& 66.9	& 68.1	& 61.0	& 64.3	& 66.4	& 60.5	& 53.9	& 68.0	& 68.2	& 59.6	& 65.4	&63.4 &{90.0} &{58.4}\\
    &  &  &  & & & $\pm$0.3 &$\pm$0.2 &$\pm$0.1 &$\pm$0.5 &$\pm$0.3 &$\pm$0.3 &$\pm$0.3 &$\pm$0.4 &$\pm$0.2 &$\pm$0.2 &$\pm$0.2 &$\pm$0.4 &$\pm$0.3 &$\pm$0.8 &$\pm$0.9\\
    OVANet + $\mathcal{L}_{tar}^{con}$ & \xmark  & \cmark & \cmark & \xmark  & \xmark & 59.5	& 67.3	& 70.1	& 61.0	& 64.5	& 67.4	& 60.1	& 52.9	& 68.4	& 68.4	& 59.2	& 66.9	& \textbf{63.9} &\textbf{91.3} &\textbf{61.8}\\
    &  &  &  & & & $\pm$0.3 &$\pm$0.2 &$\pm$0.4 &$\pm$0.2 &$\pm$0.4 &$\pm$0.3 &$\pm$0.4 &$\pm$0.4 &$\pm$0.3 &$\pm$0.4 &$\pm$0.2 &$\pm$0.1 &$\pm$0.3 &$\pm$0.7 &$\pm$0.8\\
    \midrule 
    \midrule
    Source-only &\cmark      &-       &-       &-    & -   & 46.1  & 63.3  & 72.9  & 42.8  & 54.0  & 58.7  & 47.8  & {36.1} & 66.2  & 60.8  & 45.3  & 68.2  & 55.2  & 69.6      &29.1  \\
    SHOT-O~\cite{shot}  & \cmark & \xmark  & \cmark & \xmark 
 & \xmark  & 37.7  & 41.8  & 48.4  & 56.4  & 39.8  & 40.9  & 60.0  & 41.5  & 49.7  & 61.8  & 41.4  & 43.6  & 46.9  & 77.5  & 28.1 \\
    \midrule
    UMAD~\cite{liang2021umad}  & \cmark & \cmark & \cmark & \xmark & \xmark & 53.6	 & 63.4	& 73.1	& 54.3	& 64.7	& 66.7	& 58.2	& 53.5	& 71.0	& 67.3	& 54.2	& 72.5	& 62.7 & 92.2 & 66.8\\
    &  &  &  & & & $\pm$0.8 &$\pm$0.3 &$\pm$0.1 &$\pm$0.3 &$\pm$0.1 &$\pm$0.4 &$\pm$0.5 &$\pm$0.4 &$\pm$0.1 &$\pm$0.4 &$\pm$1.0 &$\pm$0.2 &$\pm$0.4 &$\pm$0.3 &$\pm$0.1\\
    UMAD + $\mathcal{L}_{tar}^{con}$ & \cmark  & \cmark & \cmark & \xmark  & \xmark  & 56.9	& 66.8	& 73.8	& 56.7	& 65.3	& 67.2	& 59.0	& 54.9	& 71.6	& 67.0	& 57.2	& 72.9	& \textbf{64.0} &\textbf{93.3} & \textbf{67.2}\\
    &  &  &  & & & $\pm$0.5 &$\pm$0.1 &$\pm$0.2 &$\pm$0.2 &$\pm$0.3 &$\pm$0.1 &$\pm$0.2 &$\pm$0.1 &$\pm$0.3 &$\pm$0.3 &$\pm$0.5 &$\pm$0.2 &$\pm$0.3  &$\pm$0.2 &$\pm$0.2\\
    \midrule
    GLC   & \cmark & \cmark & \cmark & \cmark & \cmark & 64.7     & 72.7      & 79.1     & 60.3       &71.5     &74.2     &63.7      &63.2      &76.6     & 67.1      &64.3     &78.0     &69.6  & 90.1 & 71.5\\
    &  &  &  & & & $\pm$0.5 &$\pm$1.3 &$\pm$0.6 &$\pm$0.3 &$\pm$0.5 &$\pm$0.6 &$\pm$0.2 &$\pm$0.2 &$\pm$1.1 &$\pm$0.2 &$\pm$0.4 &$\pm$0.2 &$\pm$0.5 &$\pm$0.2 &$\pm$0.7 \\
    GLC++  & \cmark & \cmark & \cmark & \cmark & \cmark & 65.2 & 72.9 & 78.6 & 61.7 & 71.9 & 74.2 & 63.2 & 65.3 & 76.5 & 67.4 & 65.9 & 77.7 & \textbf{70.1} & \textbf{91.4} & \textbf{73.1}\\
    &  &  &  & & & $\pm$0.4 &$\pm$1.1 &$\pm$0.3 &$\pm$0.4 &$\pm$1.2 &$\pm$0.5 &$\pm$0.3 &$\pm$0.1 &$\pm$0.4 &$\pm$0.2 &$\pm$0.4 &$\pm$0.3 &$\pm$0.5 &$\pm$0.3 &$\pm$0.1\\
    \bottomrule
    \end{tabular}%
  }
  \vspace{-0.10in}
  \label{tab:osda}%
\end{table*}%
\\
\noindent \textbf{Baseline methods:} We compare GLC and GLC++ techniques with both source-available and source-free methods to empirically demonstrate the merit of our approaches. Source-available methods typically necessitate access to raw data of the source domain, whereas source-free methods operate on pre-trained models from the source domain. Remarkably, our GLC and GLC++ require only a pre-trained closed-set source model, eschewing the need for dedicated model architectures as in prior works~\cite{kundu2020universal, liang2021umad}. We benchmark our methods against a suite of existing techniques, including UAN~\cite{uan}, CMU~\cite{cmu}, ROS~\cite{ros}, DANCE~\cite{dance}, DCC~\cite{dcc}, GATE~\cite{gate}, OVANet~\cite{ova}. For a fair comparison, all these methods are performed without prior knowledge of category shift. To further demonstrate the superiority, we compare our techniques with methods tailored for specific scenarios: for OSDA, we involve STAmax~\cite{liu2019_stamax}, OSBP~\cite{osbp}, and SHOT-O~\cite{shot}; for PDA, we include PADA~\cite{pda}, ETN~\cite{ETN}, BA3US~\cite{ba3us}, and SHOT-P~\cite{shot}; and for CLDA we compare against CDAN~\cite{cdan}, MDD~\cite{MDD_ICML_19}, and SHOT~\cite{shot}. Note that the results for UMAD and OVANet are derived from our implementation, where we have also integrated our contrastive affinity learning strategy. The results for other methods are directly cited from their respective publications. The term "Source-only" denotes the baseline where the pre-trained source model is directly applied for target label prediction without any adaptation.
\\
\noindent \textbf{Implementation details:} We adopt the same network architecture as existing baseline methods. Concretely, we employ the {ResNet-50}~\cite{resnet} pre-trained on ImageNet~\cite{imagenet} as the backbone for all datasets. During the model adaptation, we leverage the SGD optimizer with a momentum of 0.9. The batch size is set to 64 for all benchmark datasets. We set the learning rate to 1e-3 for Office-31 and Office-Home, and 1e-4 for VisDA and DomainNet. For hyper-parameters, as we described in previous sections, we set $\rho$ to 0.75, $|L| = |\mathcal{X}_{pos}| = |\mathcal{X}_{neg}| = 4 $ for all benchmarks. As for $\eta$, we set it to 0.3 for Office-31, VisDA, and 1.5 for Office-Home and DomainNet. {All experiments are conducted using PyTorch 1.10 with five different random seeds \{2021, 2022, 2023, 2024, 2025\}. Notably, we do not apply any target domain augmentation (e.g., ten-crop ensemble) during evaluation.}
\begin{table*}[htbp]
  \centering
  \caption{{Accuracy (\%) comparison in PDA scenarios on the Office-Home, Office-31, and VisDA datasets, respectively.}}
  \vspace{-0.05in}
  \addtolength{\tabcolsep}{-4.0pt}
  \resizebox{0.99\textwidth}{!}{
    \begin{tabular}{lccccc|cccccccccccca|a|a}
    \toprule
    \multirow{2}[4]{*}{Methods} & \multirow{2}[4]{*}{SF} & \multirow{2}[4]{*}{OPDA} & \multirow{2}[4]{*}{OSDA} & \multirow{2}[4]{*}{PDA} & \multirow{2}[4]{*}{CLDA} & \multicolumn{13}{c|}{Office-Home} & \multicolumn{1}{c|}{Office31}  &
    \multicolumn{1}{c}{VisDA} \\
\cmidrule{7-21}    &      &       &       &       &       & Ar2Cl & Ar2Pr & Ar2Re & Cl2Ar & Cl2Pr & Cl2Re & Pr2Ar & Pr2Cl & Pr2Re & Re2Ar & Re2Cl & Re2Pr & Avg   & Avg   & \textbf{S2R} \\
    \midrule
    {PADA}~\cite{pda} & \xmark  & \xmark  & \xmark  & \cmark & \xmark & 52.0  & 67.0  & 78.7  & 52.2  & 53.8  & 59.1  & 52.6  & 43.2  & 78.8  & 73.7  & 56.6  & 77.1  & 62.1  & 92.7  & 53.5  \\
    {ETN}~\cite{ETN} & \xmark  & \xmark  & \xmark  & \cmark & \xmark & 59.2  & 77.0  & 79.5  & 62.9  & 65.7  & 75.0  & 68.3  & 55.4  & 84.4  & 75.7  & 57.7  & 84.5  & 70.4  & {{96.7}}  & 59.8  \\
    {BA3US}~\cite{ba3us} & \xmark  & \xmark  & \xmark  & \cmark & \xmark & {{60.6}}  & {{83.2}}  & {{88.4}}  & 71.8  & {{72.8}}  & 83.4  & {{75.5}}  & 61.6  & {{86.5}}  & 79.3  & 62.8  & {{86.1}}  & {{76.0}}  & {{97.8}}  & 54.9  \\
    {CMU}~\cite{cmu} & \xmark  & \cmark & \xmark  & \xmark  & \xmark & 50.9  & 74.2  & 78.4  & 62.2  & 64.1  & 72.5  & 63.5  & 47.9  & 78.3  & 72.4  & 54.7  & 78.9  & 66.5  & 83.4  & 65.5  \\
    {DANCE}~\cite{dance} & \xmark  & \cmark & \cmark & \cmark & \cmark & 53.6  & 73.2  & 84.9  & 70.8  & 67.3  & 82.6  & 70.0  & 50.9  & 84.8  & 77.0  & 55.9  & 81.8  & 71.1  & 86.0  & 73.7  \\
    {DCC}~\cite{dcc} & \xmark  & \cmark & \cmark & \cmark & \cmark & 54.2  & 47.5  & 57.5  & {{83.8}}  & 71.6  & {{86.2}}  & 63.7  & {{65.0}}  & 75.2  & {{85.5}}  & {{78.2}}  & 82.6  & 70.9  & 93.3  & 72.4  \\
    {GATE}~\cite{gate} & \xmark  & \cmark & \cmark & \cmark & \cmark & 55.8  & 75.9  & 85.3  & 73.6  & 70.2  & 83.0  & 72.1  & 59.5  & 84.7  & 79.6  & {{63.9}}  & 83.8  & 74.0  & 93.7  & {{75.6}}  \\
    \midrule
    {OVANet}~\cite{ova} & \xmark  & \cmark & \cmark & \xmark  & \xmark   & 50.9	& 71.2	& 80.2	& 60.9	& 61.1	& 69.3	& 61.1	& 45.4	& 77.6	& 72.2	& 52.7	& 77.6	& 65.0 & 89.4 &58.5\\
    &  &  &  & & & $\pm$0.5 &$\pm$0.4 &$\pm$0.2 &$\pm$1.0 &$\pm$0.5 &$\pm$0.6 &$\pm$0.4 &$\pm$0.4 &$\pm$0.2 &$\pm$0.3 &$\pm$0.5 &$\pm$0.3 &$\pm$0.4 &$\pm$0.4 &$\pm$0.9\\
    OVANet + $\mathcal{L}_{tar}^{con}$ & \xmark  & \cmark & \cmark & \xmark  & \xmark  & 55.7	& 77.2	& 82.8	& 67.1	& 67.8	& 73.2	& 70.1	& 52.5	& 80.3	& 76.3	& 58.0	& 81.3	& \textbf{70.2}  & \textbf{94.2} & \textbf{67.3}\\
    &  &  &  & & & $\pm$0.6 &$\pm$0.4 &$\pm$0.6 &$\pm$0.5 &$\pm$0.6 &$\pm$1.0 &$\pm$0.5 &$\pm$0.5 &$\pm$0.3 &$\pm$0.5 &$\pm$0.2 &$\pm$0.1 &$\pm$0.5 &$\pm$0.3  &$\pm$0.5\\
    \midrule
     \midrule
    Source-only & \cmark & -     & -     & -   & -  & 45.9  & 69.2  & 81.1  & 55.7  & 61.2  & 64.8  & 60.7  & 41.1  & 75.8  & 70.5  & 49.9  & 78.4  & 62.9  & 87.8  & 42.8  \\
    SHOT-P~\cite{shot} & \cmark & \xmark  & \xmark & \cmark & \xmark & {{64.7}}  & {{85.1}}  & {{90.1}}  & {{75.1}}  & {{73.9}}  & {{84.2}}  & {{76.4}}  & {{64.1}}  & {{90.3}}  & {{80.7}}  & 63.3  & {{85.5}}  & {{77.8}}  & 92.2  & 74.2  \\
    \midrule
    UMAD~\cite{liang2021umad}  & \cmark & \cmark & \cmark & \xmark & \xmark & 51.6     & 65.9     & 79.3     & 62.6     & 62.1     & 67.8     & 64.3     & 54.8     & 75.9     & 75.5     & 57.4     & 78.9     & 66.3 & 89.5 & 68.7\\
    &  &  &  & & & $\pm$0.5 &$\pm$0.2 &$\pm$0.4 &$\pm$0.7 &$\pm$0.7 &$\pm$0.3 &$\pm$1.2 &$\pm$0.3 &$\pm$0.7 &$\pm$0.9 &$\pm$0.8 &$\pm$0.6 &$\pm$0.6 &$\pm$0.5 &$\pm$0.2 \\
    UMAD + $\mathcal{L}_{tar}^{con}$ & \cmark  & \cmark & \cmark & \xmark  & \xmark  & 56.2	& 68.7	& 80.7	& 67.5	& 65.9	& 72.5	& 70.1	& 60.4	& 78.3	& 79.1	& 62.7	& 82.7	&\textbf{70.4} &\textbf{91.4} &\textbf{74.3}\\
    &  &  &  & & & $\pm$0.5 &$\pm$0.3 &$\pm$0.3 &$\pm$0.8 &$\pm$0.9 &$\pm$0.5 &$\pm$0.8 &$\pm$1.2 &$\pm$0.7 &$\pm$0.4 &$\pm$0.6 &$\pm$0.3 &$\pm$0.6  &$\pm$0.4 &$\pm$0.1 \\
    \midrule
    GLC   & \cmark & \cmark & \cmark & \cmark & \cmark & 56.6      & 79.0      & 87.8      & 72.0      & 71.6      & 82.7      & 74.9      & 41.3      & 82.5      & 77.4      & 59.8      & 84.3      & 72.5 & 94.0 & 76.2\\
    &  &  &  & & & $\pm$0.3 &$\pm$0.7 &$\pm$0.4 &$\pm$0.3 &$\pm$0.3 &$\pm$0.2 &$\pm$0.4 &$\pm$0.2 &$\pm$0.1 &$\pm$0.2 &$\pm$0.5 &$\pm$0.5 &$\pm$0.3  &$\pm$0.3 &$\pm$0.2\\
    GLC++  & \cmark & \cmark & \cmark & \cmark & \cmark & 58.1 & 79.6 & 89.3 & 74.8 & 72.2 & 80.6 & 76.6 & 42.0 & 83.2 & 78.5 & 60.8 & 85.2 & \textbf{73.4} & \textbf{94.1} & \textbf{75.8}\\
    &  &  &  & & & $\pm$0.3 &$\pm$0.3 &$\pm$0.3 &$\pm$0.3 &$\pm$0.3 &$\pm$0.1 &$\pm$0.4 &$\pm$0.1 &$\pm$0.4 &$\pm$0.1 &$\pm$0.2 &$\pm$0.1 &$\pm$0.2  &$\pm$0.3 &$\pm$0.4\\
    \bottomrule
    \end{tabular}%
  }
  \label{tab:pda}%
  \vspace{-0.10in}
\end{table*}%
\begin{table*}[htbp]
  \centering
  \caption{{Accuracy (\%) comparison in CLDA scenarios on the Office-Home and Office-31 datasets, respectively.}}
  \vspace{-0.05in}
  \addtolength{\tabcolsep}{-4.0pt}
  \resizebox{0.99\textwidth}{!}{
    \begin{tabular}{lccccc|cccccccccccca|a}
    \toprule
    \multirow{2}[3]{*}{Methods} & \multirow{2}[3]{*}{SF} & \multirow{2}[3]{*}{OPDA} & \multirow{2}[3]{*}{OSDA} & \multirow{2}[3]{*}{PDA}  & \multirow{2}[3]{*}{CLDA} & \multicolumn{13}{c|}{Office-Home}      & \multicolumn{1}{c}{Office-31}  \\
\cmidrule{7-20}          &       &       &       &       &       & Ar2Cl & Ar2Pr & Ar2Re & Cl2Ar & Cl2Pr & Cl2Re & Pr2Ar & Pr2Cl & Pr2Re & Re2Ar & Re2Cl & Re2Pr & Avg   & Avg \\
\midrule
    CDAN~\cite{cdan}  & \xmark  & \xmark  & \xmark  & \xmark  & \cmark & 49.0  & 69.3  & 74.5  & 54.4  & 66.0  & 68.4  & 55.6  & 48.3  & 75.9  & 68.4  & 55.4  & 80.5  & 63.8  & 86.6  \\
    MDD~\cite{MDD_ICML_19}   & \xmark  & \xmark  & \xmark  & \xmark  & \cmark & {54.9}  & 73.7  & 77.8  & 60.0  & 71.4  & 71.8  & 61.2  & 53.6  & 78.1  & 72.5  & {60.2}  & 82.3  & 68.1  & {88.9}  \\
    UAN~\cite{uan}   & \xmark  & \cmark & \xmark  & \xmark  & \xmark  & 45.0  & 63.6  & 71.2  & 51.4  & 58.2  & 63.2  & 52.6  & 40.9  & 71.0  & 63.3  & 48.2  & 75.4  & 58.7  & 84.4  \\
    CMU~\cite{cmu}   & \xmark  & \cmark & \xmark  & \xmark  & \xmark  & 42.8  & 65.6  & 74.3  & 58.1  & 63.1  & 67.4  & 54.2  & 41.2  & 73.8  & 66.9  & 48.0  & 78.7  & 61.2  & 79.9  \\
    DANCE~\cite{dance} & \xmark  & \cmark & \cmark & \cmark & \cmark  & 54.3  & 75.9  & 78.4  & 64.8  & 72.1  & 73.4  & 63.2  & 53.0  & 79.4  & {73.0}  & 58.2  & 82.9  & 69.1  & 85.5  \\
    DCC~\cite{dcc}   & \xmark  & \cmark & \cmark & \cmark & \cmark  & 35.4  & 61.4  & 75.2  & 45.7  & 59.1  & 62.7  & 43.9  & 30.9  & 70.2  & 57.8  & 41.0  & 77.9  & 55.1  & 87.4  \\
    GATE~\cite{gate} & \xmark  & \cmark & \cmark & \cmark & \cmark  & 54.6 & 76.9 & 79.8 & 66.1 & 73.5 & 74.2 & 65.3 & 54.8 & 80.6 & 73.9 & 59.5 & 83.7 & 70.2 & 88.3\\
    \midrule
    OVANet~\cite{ova} & \xmark  & \cmark & \cmark & \xmark  & \xmark & 48.6	& 68.6	& 76.1	& 58.2	& 66.0	& 68.9	& 55.9	& 44.7	& 75.9	& 68.9	& 52.0	& 79.8	& 63.6 & 82.6 \\
    &  &  &  & & & $\pm$0.3 &$\pm$0.1 &$\pm$0.1 &$\pm$0.6 &$\pm$0.1 &$\pm$0.5 &$\pm$0.5 &$\pm$0.1 &$\pm$0.1 &$\pm$0.2 &$\pm$0.4 &$\pm$0.1 &$\pm$0.3 &$\pm$0.3\\
    OVANet + $\mathcal{L}_{tar}^{con}$ & \xmark  & \cmark & \cmark & \xmark  & \xmark  & 51.8	& 72.3	& 77.0	& 61.1	& 68.5	& 70.1	& 60.8	& 48.5	& 77.2	& 71.1	& 54.7	& 81.4	& \textbf{66.2} &\textbf{86.6} \\
    &  &  &  & & & $\pm$0.2 &$\pm$0.1 &$\pm$0.2 &$\pm$0.2 &$\pm$0.3 &$\pm$0.3 &$\pm$0.2 &$\pm$0.2 &$\pm$0.2 &$\pm$0.4 &$\pm$0.3 &$\pm$0.3 &$\pm$0.2 &$\pm$0.3\\
    \midrule
    \midrule
    Source-only & \cmark & -     & -     & -     & -     & 44.8  & 67.4  & 74.2  & 53.0  & 63.3  & 65.1  & 53.7  & 40.5  & 73.5  & 65.6  & 46.3  & 78.3  & 60.5  & 78.8 \\
    \midrule
    SHOT~\cite{shot} &\cmark &\xmark &\xmark &\xmark &\cmark & 56.1 & 77.6 & 80.6 & 68.2 & 79.5	& 78.4	& 68.0	& 54.8	& 81.9	& 73.2	& 60.4 & 84.0	& 71.9 & 88.2 \\
    &  &  &  & & & $\pm$0.3 &$\pm$0.1 &$\pm$0.2 &$\pm$0.2 &$\pm$0.3 &$\pm$0.4 &$\pm$0.2 &$\pm$0.2 &$\pm$0.1 &$\pm$0.1 &$\pm$0.1 &$\pm$0.2 &$\pm$0.2 &$\pm$0.2 \\
    SHOT + $\mathcal{L}_{tar}^{con}$ &\cmark &\xmark &\xmark &\xmark &\cmark & 57.3	& 78.2	& 80.9	& 68.4	& 80.2	& 78.5	& 68.5	& 55.7	& 82.1	& 73.9	& 61.6	& 84.4	& \textbf{72.5} &\textbf{89.0}\\
    &  &  &  & & & $\pm$0.2 &$\pm$0.1 &$\pm$0.1 &$\pm$0.3 &$\pm$0.2 &$\pm$0.3 &$\pm$0.2 &$\pm$0.1 &$\pm$0.1 &$\pm$0.2 &$\pm$0.1 &$\pm$0.2 &$\pm$0.2 &$\pm$0.2 \\
    \midrule
    UMAD~\cite{liang2021umad}  & \cmark & \cmark & \cmark & \xmark  & \xmark  & 47.7  & 64.6  & 73.0  & 58.8  & 64.7  & 68.1  & 58.2  & 47.4  & 74.2  & 69.2  & 52.7  & 77.9  & 63.1  & 81.4\\
    &  &  &  & & & $\pm$0.3 &$\pm$0.3 &$\pm$0.2 &$\pm$0.1 &$\pm$1.0 &$\pm$0.5 &$\pm$0.1 &$\pm$0.2 &$\pm$0.1 &$\pm$0.3 &$\pm$0.1 &$\pm$0.2 &$\pm$0.3 &$\pm$0.3\\
    UMAD + $\mathcal{L}_{tar}^{con}$ & \cmark  & \cmark & \cmark & \xmark  & \xmark  & 49.2	& 66.4	& 73.5	& 59.3	& 66.4	& 68.9	& 59.2	& 48.6	& 75.0	& 69.6	& 54.3	& 78.7	& \textbf{64.1} & \textbf{82.7}\\
    &  &  &  & & & $\pm$0.3 &$\pm$0.3 &$\pm$0.2 &$\pm$0.3 &$\pm$0.5 &$\pm$0.5 &$\pm$0.2 &$\pm$0.3 &$\pm$0.1 &$\pm$0.1 &$\pm$0.3 &$\pm$0.2 &$\pm$0.3 &$\pm$0.2\\
    \midrule
    GLC   & \cmark & \cmark & \cmark & \cmark & \cmark  & 52.5  & {76.6}  & {80.2}  & {65.2}  & {78.5}  & {78.3}  & {64.8}  & {55.1}  & {81.5}  & 70.9  & 58.1  & {83.8}  & {70.5}  & {88.0}\\
    &  &  &  & & & $\pm$0.5 &$\pm$0.5 &$\pm$0.2 &$\pm$0.1 &$\pm$0.2 &$\pm$0.2 &$\pm$0.2 &$\pm$0.6 &$\pm$0.2 &$\pm$0.3 &$\pm$0.2 &$\pm$0.2 &$\pm$0.3  &$\pm$0.2\\
    GLC++  & \cmark & \cmark & \cmark & \cmark & \cmark & 53.5 & 77.0 & 80.3 & 65.7 & 79.6 & 78.7 & 65.6 & 56.1 & 81.6 & 71.5 & 58.5 & 84.0 & \textbf{71.0} & \textbf{88.4} \\
    &  &  &  & & & $\pm$0.4 &$\pm$0.1 &$\pm$0.1 &$\pm$0.1 &$\pm$0.4 &$\pm$0.1 &$\pm$0.3 &$\pm$0.2 &$\pm$0.1 &$\pm$0.3 &$\pm$0.1 &$\pm$0.1  &$\pm$0.2 &$\pm$0.2\\
    \bottomrule
    \end{tabular}%
    }
  \label{tab:clda}%
  \vspace{-0.15in}
\end{table*}%

\subsection{Experiment Results}
\noindent \textbf{Results for OPDA:} We first conduct experiments in the most challenging setting, i.e., OPDA, in which both source and target domains involve private categories. Results on the Office-Home dataset are summarized in Table~\ref{tab:opda_Office-Home}, and results on the Office-31, VisDA, and DomainNet datasets are detailed in Table~\ref{tab:opda_rest}. As shown in Table~\ref{tab:opda_Office-Home} and Table~\ref{tab:opda_rest}, our GLC achieves new state-of-the-art, outperforming not only source-free but also source-available methods across these datasets. Especially, on VisDA, GLC achieves an H-score of 73.2 ± 0.4\%, significantly surpassing GATE~\cite{gate} and UMAD~\cite{liang2021umad} by 16.8\% and 18.1\%, respectively. Similarly, on DomainNet, the largest benchmark for domain adaptation, GLC still demonstrates substantial performance improvement over UMAD and GATE, with gains of approximately 10.4\% and 2.9\%. The starting point of our contrastive affinity learning strategy is to enhance the discrimination of different "unknown" clusters, the results in Table~\ref{tab:opda_Office-Home} and Table~\ref{tab:opda_rest} demonstrate that with this strategy, GLC++ further amplifies H-score results across the Office-31, Office-Home, and VisDA datasets. 
For instance, GLC++ improves upon GLC on VisDA by 2.1\% H-score and achieves a 6.6\% increase in performance in the Ar$\rightarrow$Cl task on the Office-Home dataset. Nevertheless, we observe a marginal decrease of 0.2\% on DomainNet. This decrease could be attributed to gradient conflicts arising from a sub-optimal weighting of training loss, leading to the optimization process inadvertently compromising one training objective in favor of improving the overall objective. Similar performance enhancements are attained when integrating the contrastive affinity learning strategy with other methods, such as OVANet and UMAD. In particular, this strategy increases the H-score of OVANet/UMAD from 48.1\%/55.1\% to 53.4\%/56.8\% on VisDA, respectively. \\
\noindent \textbf{Results for OSDA:} We then conduct experiments in OSDA scenarios, where only the target domain involves categories not presented in the source domain. Results on the Office-Home, Office-31, and VisDA datasets are presented in Table~\ref{tab:osda}. GLC achieves results surpassing or comparable to the state-of-the-art. Specifically, GLC obtains an H-score of 69.6 ± 0.5\% on Office-Home and 71.5 ± 0.7\% on VisDA, with an improvement of 6.9\% and 4.7\% compared to UMAD, respectively. Further improvements are observed with GLC++, which boosts up the H-score to 70.1 ± 0.5\% on Office-Home, from 90.1 ± 0.2\% to 91.4 ± 0.3\% on Office-31, and 71.5 ± 0.7\% to 73.1 ± 0.1\% on VisDA. Moreover, the contrastive learning strategy contributes an H-score increase of 0.5\% and 1.3\% to OVANet and UMAD on Office-Home, which affirms the merit of this strategy serving as a beneficial complement to existing methods.\\
\noindent \textbf{Results for PDA:} Subsequently, we verify the effectiveness of our technique in PDA scenarios, where the label space of the target domain is a subset of the source domain. Results summarized in Table~\ref{tab:pda} show that GLC achieves better results than SHOT-P~\cite{shot} on the Office31 and VisDA datasets, despite SHOT-P being specifically tailored for PDA settings. In a fairer comparison, GLC significantly outperforms UMAD, achieving performance gains of 6.2\%, 4.5\%, and 7.5\% on the Office-31, Office-Home, and VisDA datasets, respectively. Moreover, against existing source-available methods, GLC still achieves surpassing or comparable performance. Consistent with observations in OPDA and OSDA scenarios, the integration of our contrastive affinity learning strategy further exhibits performance improvement to GLC and other baseline methods in PDA scenarios. In particular, it propels OVANet and UMAD's accuracies by substantial margins of 6.1\% and 3.8\%, respectively. \\
\noindent \textbf{Results for CLDA:}
Most existing methods tailored for category shifts are inadequately equipped for vanilla CLDA scenarios, where the label spaces across the source and target domains remain unchanged. To verify the effectiveness and versatility of GLC in CLDA scenarios, we further conduct experiments on the Office-31 and Office-Home datasets. Table~\ref{tab:clda} details the results, demonstrating the superiority of GLC against existing baselines. Specifically, GLC attains accuracies of 70.5 ± 0.3\% on Office-Home and 88.0 ± 0.2\% on Office-31, yielding significant improvements over UMAD by 7.4\% and 6.6\%, respectively. 
These results empirically confirm the broad applicability and effectiveness of GLC across diverse domain adaptation scenarios. The incorporation of our contrastive affinity learning strategy also reveals a notable performance contribution in this scenario. Specifically, it facilitates an accuracy increase of 2.6\% for OVANet, 0.6\% and 1.0\% for SHOT and UMAD on Office-Home, correspondingly. \\
\noindent \textbf{Results for Novel Category Discovery:} Prevailing approaches designed for OPDA and OSDA have made great progress in recognizing "known" data and simultaneously rejecting "unknown"  data. However, these methods typically treat the "unknown" data as a monolithic entity, overlooking the nuanced semantic distinctions within. This oversight results in sub-optimal performance in identifying diverse categories or clusters involved in "unknown" data. Drawing inspiration from recent progress in novel category discovery~\cite{han2019_dtc, hsu2019_mcl, vaze2022_gcd}, we further examine the adapted target model's capability to discern distinct categories among "unknown" data. We adopt a similar protocol like~\cite{han2019_dtc, vaze2022_gcd}, performing k-means clustering on "unknown" data with the actual count of the target private category. Table~\ref{tab:novel_category_discovery} lists the performance comparison in OPDA and OSDA scenarios. An observation is that the source-available method OVANet typically exhibits better performance in terms of novel category discovery than source-free methods, i.e., UMAD and GLC. This may be because lacking ground-truth supervision from source raw data makes it more difficult to identify different semantics contained in "unknown" data. However, the incorporation of our contrastive affinity learning strategy significantly mitigates this challenge. In particular, GLC++ elevates the NCD Accuracy of GLC from 65.2\%/50.4\% to 70.6\%/55.4\% on Office-31 and Office-Home in OPDA scenarios, and from 53.6\%/66.9\% to 57.7\%/70.8\% on Office-Home and VisDA in OSDA scenarios, respectively. This strategy is also complementary to UMAD and OVANet, delivering an improvement of 3.3\% and 1.4\% in OPDA scenarios, a boost of 4.5\% and 1.7\% boost in OSDA scenarios, across Office-31, Office-Home, and DomainNet, respectively. 
\par {It is worth noting that fluctuations in NCD Acc results are more pronounced compared to those observed in Table 2 to 6. We hypothesize that this is due to the absence of clear class pseudo-label guidance for the "unknown" data in the target domain during training, unlike the known class data. Additionally, since the NCD Acc metric is based on feature K-means clustering and does not rely on a fixed classifier for the "unknown" data, the results exhibit greater variability. However, even with these challenges, our contrastive affinity learning strategy still notably enhances the model’s ability to distinguish between different "unknown" data points.}

\begin{table}[tbp]
\centering
\caption{{Novel Category Discovery Accuracy (\%) in OPDA and OSDA scenarios.}}
\vspace{-0.10in}
\addtolength{\tabcolsep}{-1.0pt}
\resizebox{0.48\textwidth}{!}{
\begin{tabular}{clccc}
\toprule
Setting               & Methods   & Office-31 & Office-Home & DomainNet  \\
\midrule
\multirow{9}{*}{OPDA} & OVANet    & 69.4$\pm$2.6       & 51.6$\pm$1.4   & 43.5$\pm$3.0        \\
                      & OVANet + $\mathcal{L}_{tar}^{con}$ & 70.5$\pm$2.2    & 53.5$\pm$1.2          & 44.7 $\pm$2.6      \\
                      & &\textcolor{my_darkgreen}{\textbf{+1.1}}
                      &\textcolor{my_darkgreen}{\textbf{+1.9}}
                      &\textcolor{my_darkgreen}{\textbf{+1.2}}\\
                      \cmidrule{2-5}
                      & UMAD      &  61.8$\pm$3.2         & 44.6$\pm$1.8   & 33.2$\pm$2.7         \\
                      &UMAD +$\mathcal{L}_{tar}^{con}$   & 67.4$\pm$2.5    & 48.0$\pm$1.5   & 34.1$\pm$2.3          \\
                      & &\textcolor{my_darkgreen}{\textbf{+5.6}}
                      &\textcolor{my_darkgreen}{\textbf{+3.4}}
                      &\textcolor{my_darkgreen}{\textbf{+0.9}}\\
                      \cmidrule{2-5}
                      & GLC       & 65.2$\pm$2.6        & 50.4$\pm$1.4          & 39.4$\pm$2.1       \\
                      & GLC++     & 70.6$\pm$2.1        & 55.4$\pm$1.2          & 40.1$\pm$1.7        \\
                      & &\textcolor{my_darkgreen}{\textbf{+5.4}}
                      &\textcolor{my_darkgreen}{\textbf{+5.0}}
                      &\textcolor{my_darkgreen}{\textbf{+0.7}}\\
\midrule
\midrule
Setting               & Methods   & Office-31 & Office-Home & VisDA  \\
\midrule
\multirow{9}{*}{OSDA} & OVANet    & 71.0$\pm$2.4         & 54.2$\pm$1.3     & 68.6$\pm$0.9         \\
                      & OVANet +$\mathcal{L}_{tar}^{con}$ & 73.3$\pm$2.3    & 55.3$\pm$1.4    & 70.4$\pm$1.4         \\
                      & &\textcolor{my_darkgreen}{\textbf{+2.3}}
                      &\textcolor{my_darkgreen}{\textbf{+1.1}}
                      &\textcolor{my_darkgreen}{\textbf{+1.8}}\\
                      \cmidrule{2-5}
                      & UMAD      & 62.8$\pm$2.6         & 46.6$\pm$1.4          & 56.9$\pm$1.3         \\
                      & UMAD  +$\mathcal{L}_{tar}^{con}$   & 68.7$\pm$2.0      & 49.7$\pm$1.4  & 61.5$\pm$1.5         \\
                      & &\textcolor{my_darkgreen}{\textbf{+5.9}}
                      &\textcolor{my_darkgreen}{\textbf{+3.1}}
                      &\textcolor{my_darkgreen}{\textbf{+4.6}}\\
                      \cmidrule{2-5}
                      & GLC       & 67.4$\pm$2.7          & 53.6$\pm$1.7          & 66.9$\pm$1.4    \\
                      & GLC++     & 70.3$\pm$2.5         & 57.7$\pm$1.3          & 70.8$\pm$1.1        \\
                      & &\textcolor{my_darkgreen}{\textbf{+2.9}}
                      &\textcolor{my_darkgreen}{\textbf{+4.1}}
                      &\textcolor{my_darkgreen}{\textbf{+3.9}}\\
\bottomrule
\end{tabular}
\label{tab:novel_category_discovery}
}
\vspace{-0.10in}
\end{table}

\begin{table}[tbp]
  \centering
  \caption{{Ablation Study}. H-score (\%) in OPDA scenarios on Office-31, Office-Home, and VisDA, with different variants of GLC and GLC++.}
    \vspace{-0.10in}
  \addtolength{\tabcolsep}{0.0pt}
  \resizebox{0.48\textwidth}{!}{
    \begin{tabular}{ccc|ccc}
    \toprule
    $\mathcal{L}^{glb}_{tar}$ & $\mathcal{L}^{loc}_{tar}$ &$\mathcal{L}^{con}_{tar}$ & Office-31 & Office-Home & VisDA \\
    \midrule
    - & - & -  & 64.9  & 60.9  & 25.7 \\
    \midrule
    \cmark & - & - & 86.3$\pm$0.3  & 74.6$\pm$0.6  & 66.2$\pm$0.3 \\
    - & \cmark & -  & 87.4$\pm$0.2  & 67.3$\pm$0.5  & 57.5$\pm$0.3 \\
    \cmark & \cmark & -  & {87.6$\pm$0.3}  & {75.7$\pm$0.7}  &{73.2$\pm$0.4} \\
    \cmark & - & \cmark & \textbf{88.4$\pm$0.3} & 76.3$\pm$0.6 & 68.9$\pm$0.3 \\
    \cmark & \cmark & \cmark & {88.2$\pm$0.4} & \textbf{77.0$\pm$0.6} & \textbf{75.3$\pm$0.3}\\
    \bottomrule
    \end{tabular}%
  \label{tab:ablation}%
  }
  \vspace{-0.10in}
\end{table}%

\begin{figure}[t]
  \centering
  \includegraphics[width=0.40\textwidth]{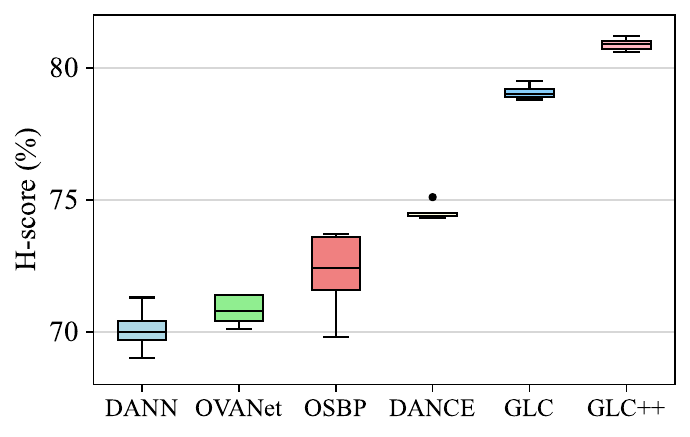}
  \vspace{-0.1in}
  \caption{{Performance on wild animal classification from virtual images to real-world images in open-set domain adaptation. H-scores (\%) of DANN, OVANet, OSBP, DANCE, GLC, and GLC++. Each boxplot shows the upper and lower quartiles, with the median represented by the horizontal line, and whiskers extending to 1.5 times the interquartile range.}}
  \label{fig:osda_animal}
\end{figure}

\begin{figure}[ht]
  \centering
  \vspace{-0.1in}
  \includegraphics[width=0.48\textwidth]{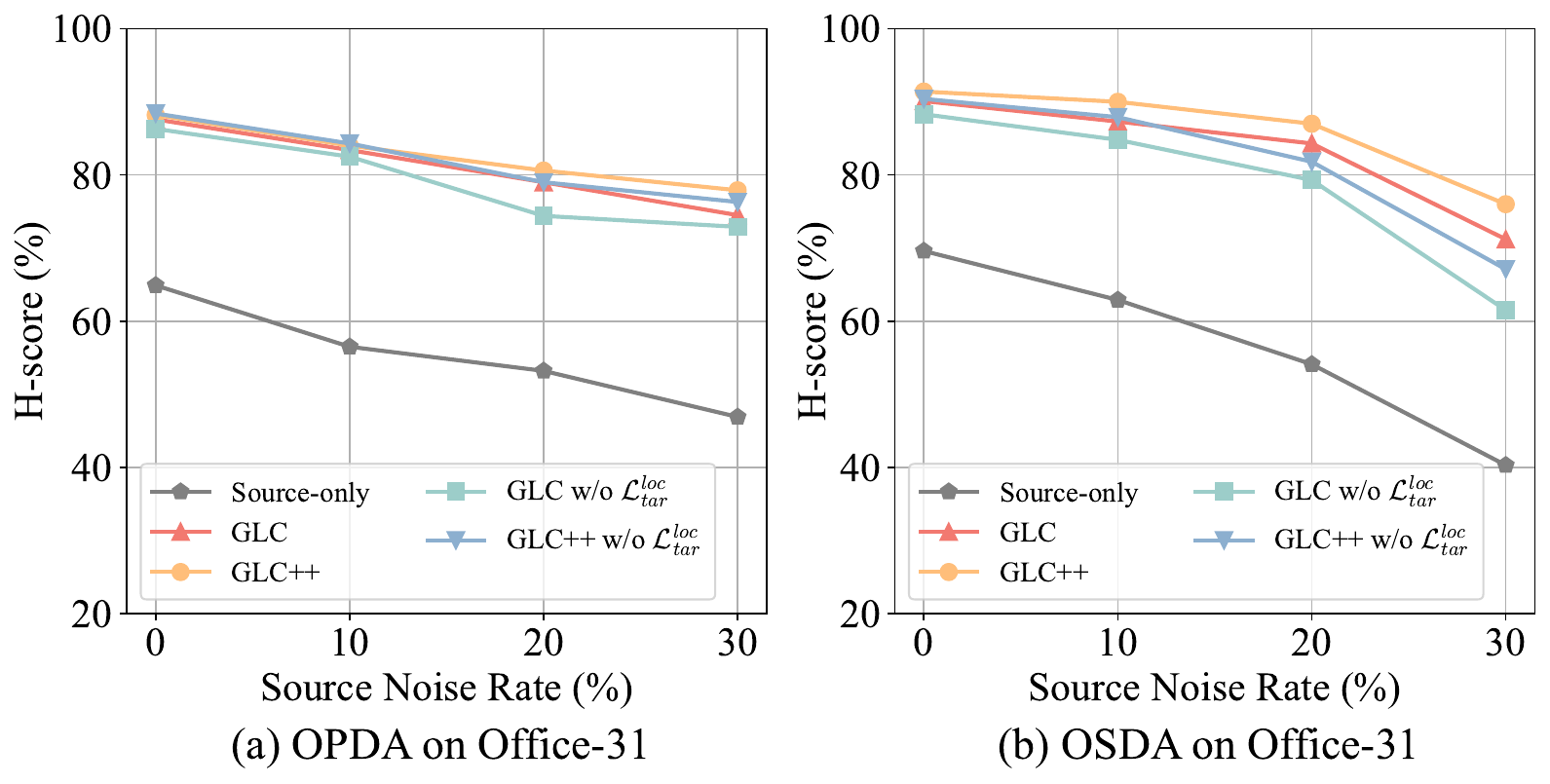}
  \vspace{-0.1in}
  \caption{H-score curves on Office-31 dataset in OPDA and OSDA settings, under varying levels of instance-independent label noise in the source data. Despite degradation in source model performance due to noise, the target models adapted by GLC and GLC++ remain robust and superior.}
  \vspace{-0.15in}
  \label{fig:src_noise}
\end{figure}

\begin{figure*}
  \centering
  \vspace{-0.1in}
  \includegraphics[width=0.99\linewidth]{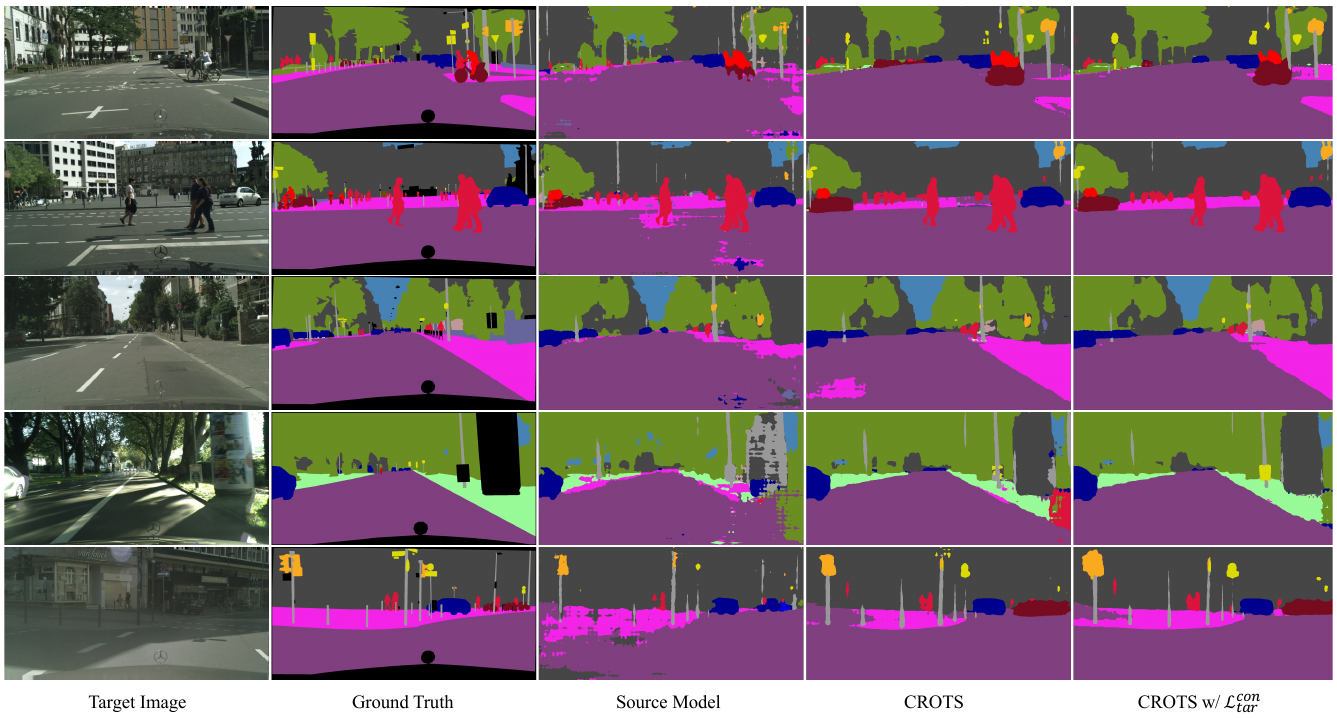}
  \vspace{-0.1in}
  \caption{{Visualization for predicted segmentation masks on the GTA5$\rightarrow$Cityscapes adaptation task.}}
  \label{fig:segmentation}
\end{figure*}

\begin{figure*}[ht]
    \centering
    \includegraphics[width=0.99\textwidth]{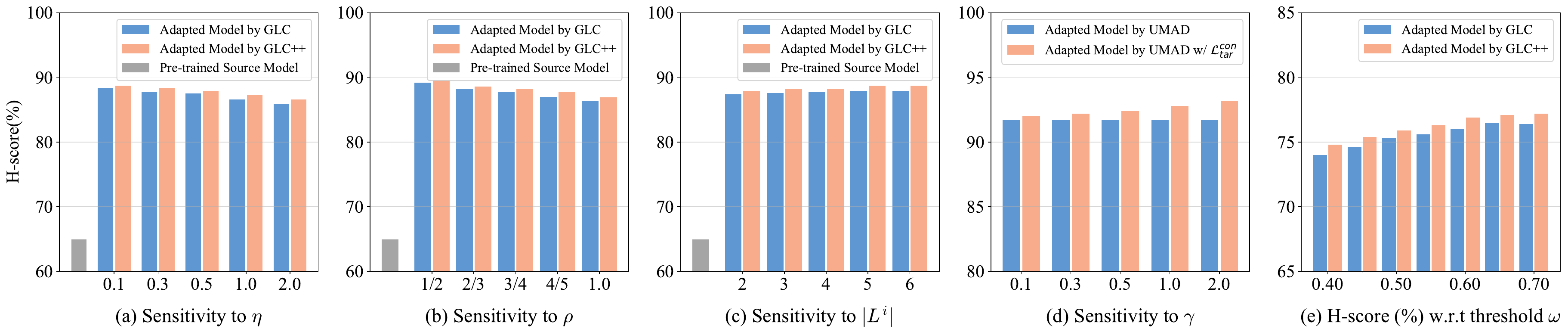}
    \vspace{-0.10in}
    \caption{{Analysis of GLC and GLC++. (a-d) present the hyper-parameter sensitivity of $\eta$, $\rho$, $|L^i|$, $\gamma$ on Office-31 in the OPDA scenario. (e) illustrates the H-score with respect to the threshold $\omega$ on Office-Home in the OPDA scenario. It is obvious that the overall results are stable around the selected hyper-parameters. We may find better parameter configurations for each setting through an oracle validation.}}
    \vspace{-0.10in}
    \label{fig:hyper_param}
\end{figure*}

\begin{table*}[t]
  \centering
  \caption{{Performance comparison of GTA-5$\rightarrow$Cityscapes adaptation in terms of mIoU (\%).}}
  \vspace{-0.05in}
  \addtolength{\tabcolsep}{-3.0pt}
  \resizebox{0.99\textwidth}{!}{
    \begin{tabular}{lcccccccccccccccccccca}
      \toprule
      Methods & SF & Road & SW & Build & Wall & Fence & Pole & TL & TS & Veg. & Terrain & Sky & PR & Rider & Car & Truck & Bus & Train & Motor & Bike & mIoU \\ 
      \midrule
      AdvEnt~\cite{vu2019_advent} & \xmark & 89.4 &33.1 &81.0 &26.6 &26.8 &27.2 &33.5 &24.7 &83.9 &36.7 &78.8 &58.7 &30.5 &84.8 &38.5 &44.5 &1.7 &31.6 &32.4 &45.5\\
      CRST~\cite{zou2019_CRST} & \xmark &91.0 &55.4 &80.0 &33.7 &21.4 &37.3 &32.9 &24.5 &85.0 &34.1 &80.8 &57.7 &24.6 &84.1 &27.8 &30.1 &26.9 &26.0 &42.3 &47.1\\
      FADA~\cite{wang2020_FADA} & \xmark &91.0 &50.6 &86.0 &43.4 &29.8 &36.8 &43.4 &25.0 &86.8 &38.3 &87.4 &64.0 &38.0 &85.2 &31.6 &46.1 &6.5 &25.4 &37.1 &50.1\\
      ProDA~\cite{zhang2021_proda} & \xmark &87.8 &56.0 &79.7 &46.3 &44.8 &45.6 &53.5 &53.5 &88.6 &45.2 &82.1 &70.7 &39.2 &88.8 &45.5 &59.4 &1.0 &48.9 &56.4 &57.5\\
      \midrule
      URMA~\cite{fleuret2021_URMA} &\cmark &92.3 &55.2 &81.6 &30.8 &18.8 &37.1 &17.7 &12.1 &84.2 &35.9 &83.8 &57.7 &24.1 &81.7 &27.5 &44.3 &6.9 &24.1 &40.4 &45.1\\
      SRDA~\cite{bateson2020_SRDA} &\cmark &90.5 &47.1 &82.8 &32.8 &28.0 &29.9 &35.9 &34.8 &83.3 &39.7 &76.1 &57.3 &23.6 &79.5 &30.7 &40.2 &0.0 &26.6 &30.9 &45.8\\
      GtA~\cite{kundu2021_GtA}   &\cmark &91.7 &53.4 &86.1 &37.6 &32.1 &37.4 &38.2 &35.6 &86.7 &48.5 &89.9 &62.6 &34.3 &87.2 &51.0 &50.8 &4.2 &42.7 &53.9 &53.4\\
      ATP~\cite{wang2024_ATP}   &\cmark &93.2 &55.8 &86.5 &45.2 &27.3 &36.6 &42.8 &37.9 &86.0 &43.1 &87.9 &63.5 &15.3 &85.5 &41.2 &55.7 &0.0 &38.1 &57.4 &52.6\\
      CROTS~\cite{luo2024_CROTS} &\cmark &94.4 &61.2 &85.5 &25.6 &31.3 &34.5 &41.9 &55.2 &87.8 &48.5 &89.7 &63.6 &32.8 &87.6 &51.2 &46.7 &0.0 &35.6 &47.5 &53.7\\
      &  &$\pm$0.4 &$\pm$0.3 &$\pm$0.1 &$\pm$0.3 &$\pm$0.5 &$\pm$0.2 &$\pm$0.2 &$\pm$0.2 &$\pm$0.1 &$\pm$0.6 &$\pm$0.6 &$\pm$0.4 &$\pm$0.3 &$\pm$0.5 &$\pm$0.6 &$\pm$0.2 &$\pm$0.0 &$\pm$0.3 &$\pm$0.3 &$\pm$0.3\\
      CROTS + $\mathcal{L}_{tar}^{con}$ & \cmark &95.2 &67.5 &85.5 &34.7 &34.7 &33.2 &42.9 &54.0 &88.0 &51.3 &89.7 &62.0 &28.6 &87.2 &54.3 &54.0 &0.0 &33.0 &51.0 &{55.1} \\
      &  &$\pm$0.3 &$\pm$0.6 &$\pm$0.1 &$\pm$0.3 &$\pm$0.5 &$\pm$0.4 &$\pm$0.2 &$\pm$0.3 &$\pm$0.1 &$\pm$0.7 &$\pm$0.5 &$\pm$0.2 &$\pm$0.2 &$\pm$0.5 &$\pm$0.6 &$\pm$0.2 &$\pm$0.0 &$\pm$0.4 &$\pm$0.2 &$\pm$0.3\\
      \bottomrule
    \end{tabular}
    \label{tab:segmentation}
  }
\end{table*}

\begin{figure*}[ht]
    \centering
    \includegraphics[width=0.99\textwidth]{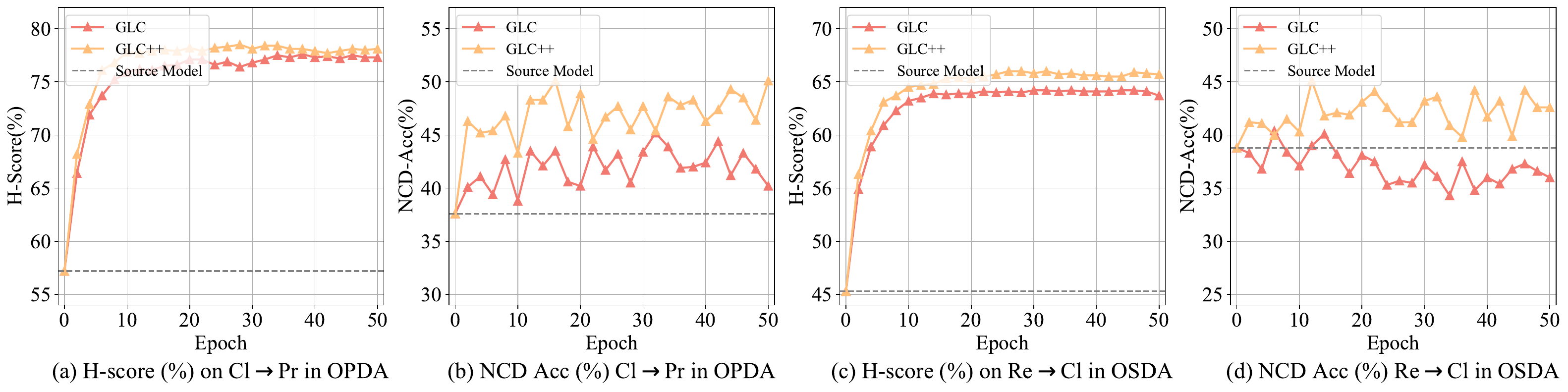}
    \vspace{-0.10in}
    \caption{The H-score and NCD Acc curves for Cl$\rightarrow$Pr in OPDA on Office-Home, and Re$\rightarrow$Cl in OSDA on Office-Home during training, respectively. GLC and GLC++ exhibit stable training convergence with respect to the H-score. Incorporating our contrastive affinity learning strategy, GLC++ consistently improves NCD Acc over the pre-trained source model, while GLC may show performance degradation in some settings.}
    \vspace{-0.10in}
    \label{fig:convergence}
\end{figure*}

\begin{figure*}[ht]
    \centering
    \includegraphics[width=0.99\textwidth]{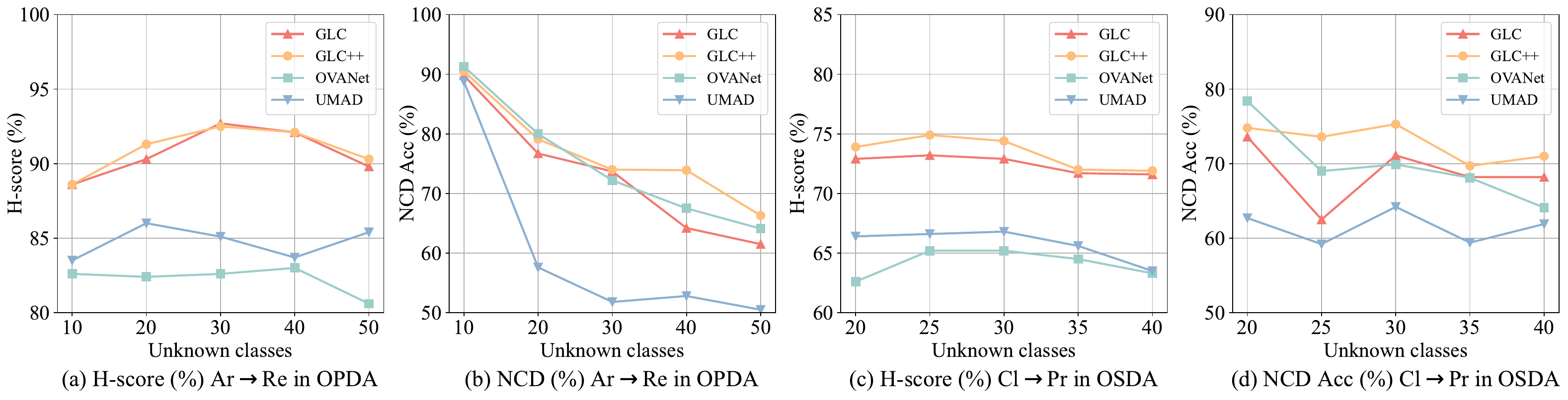}
    \vspace{-0.10in}
    \caption{{The H-score and NCD Acc curves for Ar$\rightarrow$Re in OPDA, and Cl$\rightarrow$Pr in OSDA on Office-Home,
    when varying the number of "unknown" classes. GLC and GLC++ demonstrate robust and superior H-score performance compared to existing methods. A common trend observed is the decline in NCD Acc with an increase in "unknown" classes across all baselines. The source-available method, OVANet, often outperforms source-free GLC and UMAD. We speculate that this is due to the lack of supervision from source data, making it prone to confuse different "unknown" categories. Nevertheless, with the assistance of contrastive affinity learning, GLC++ obtains comparable or surpassing NCD Acc against OVANet.}}
    \vspace{-0.10in}
    \label{fig:vary_unk}
\end{figure*}
\subsection{{Experiment beyond Standard Object Recognition Benchmarks}}
\noindent {\textbf{Results for Cross-Domain Segmentation:} Similar to object recognition tasks, semantic segmentation models also face significant performance degradation when there is a change in scenarios. Therefore, model adaptation for semantic segmentation is also crucial. 
To verify the versatility of our contrastive affinity learning strategy, we remold and incorporate it into a recent source-free domain adaptation method for semantic segmentation, CROTS \mbox{\cite{luo2024_CROTS}} ($\gamma$ is set to 0.1). We select the widely-used GTA-5 \mbox{\cite{richter2016_GTA5}} $\rightarrow$ Cityscapes \mbox{\cite{cordts2016_cityscapes}} datasets and conduct experiments. The results are summarized in Table \mbox{\ref{tab:segmentation}}. Similar to the improvements observed in image recognition, our contrastive affinity learning strategy also leads to performance gains in semantic segmentation. Specifically, it contributes an mIoU improvement of 1.5\%, demonstrating the effectiveness and versatility of our approach. Fig. \mbox{\ref{fig:segmentation}} visualizes examples for cross-domain semantic segmentation, which indicates that our contrastive affinity learning strategy can enhance the CROTS \mbox{\cite{luo2024_CROTS}} to achieve more precise and reliable results with fewer spurious areas.}

\noindent {\textbf{Results for Open-set Wild-animal Classification:} One may wonder whether GLC and GLC++ are applicable to open-world applications. To find the answer, we perform experiments on a real-world, open-set application: wild-animal classification. Accurately classifying wild animals is crucial for biodiversity research and ecosystem protection \mbox{\cite{wild_animal_1}}, especially for identifying novel species \mbox{\cite{wild_animal_2}}. Specifically, we leverage the I2AWA \mbox{\cite{I2AWA}} benchmark, which consists of a virtual source domain and a real-world target domain, both with 50 animal categories. The source domain contains 2,970 images collected from Google Image search, while the target domain includes 37,322 images from the AWA2 \mbox{\cite{AWA}} dataset. We divide the 50 categories into known (30) and unknown (20) categories. Our results, summarized in Fig. \mbox{\ref{fig:osda_animal}}, demonstrate that our GLC and GLC++ methods achieve an overall H-score of $79.1 \pm 0.3\%$ and $80.9 \pm 0.3\%$, respectively. In contrast, the baseline methods DANN, OVANet, OSBP, and DANCE achieve H-scores of $70.1 \pm 0.9\%$, $70.8 \pm 0.5\%$, $72.2 \pm 1.6\%$, and $74.5 \pm 0.3\%$, respectively. These results further validate the effectiveness of our methods in real-world applications.}

\subsection{Experiment Analysis}
\par \textbf{Ablation Study:} To thoroughly assess the individual contribution of the components comprising our methods, we conduct extensive ablation studies on the Office-31, Office-Home, and VisDA datasets in OPDA scenarios. The results are consolidated in Table~\ref{tab:ablation}. Here, $\mathcal{L}_{tar}^{glb}$, $\mathcal{L}_{tar}^{loc}$, $\mathcal{L}_{tar}^{con}$ refer to the objectives corresponding to our one-vs-all global clustering strategy, local k-NN consensus clustering strategy, and contrastive affinity learning strategy, respectively. From these results, we can conclude that these three components are not only efficacious on their own but also exhibit complementary effects to each other. In particular, the global clustering is of vital importance to help us distinguish "known" and "unknown" categories. For instance, on VisDA, the application of $\mathcal{L}_{tar}^{glb}$ alone propels the source model's H-score from 25.7\% to 66.2\%, significantly outperforming current state-of-the-art GATE by 9.8\%. With the incorporation of local k-NN strategy, it achieves remarkable performance improvement (73.2\% vs 66.2\%), basically underscoring the efficacy of local clustering strategy in alleviating negative transfer. As aforementioned, even though the starting point of contrastive affinity learning is to boost the model's capacity for novel category discovery among "unknown" data, it is also beneficial for general categorical discrimination. The ablation study further verifies this merit.
\begin{figure*}[ht]
    \centering
    \includegraphics[width=0.99\textwidth]{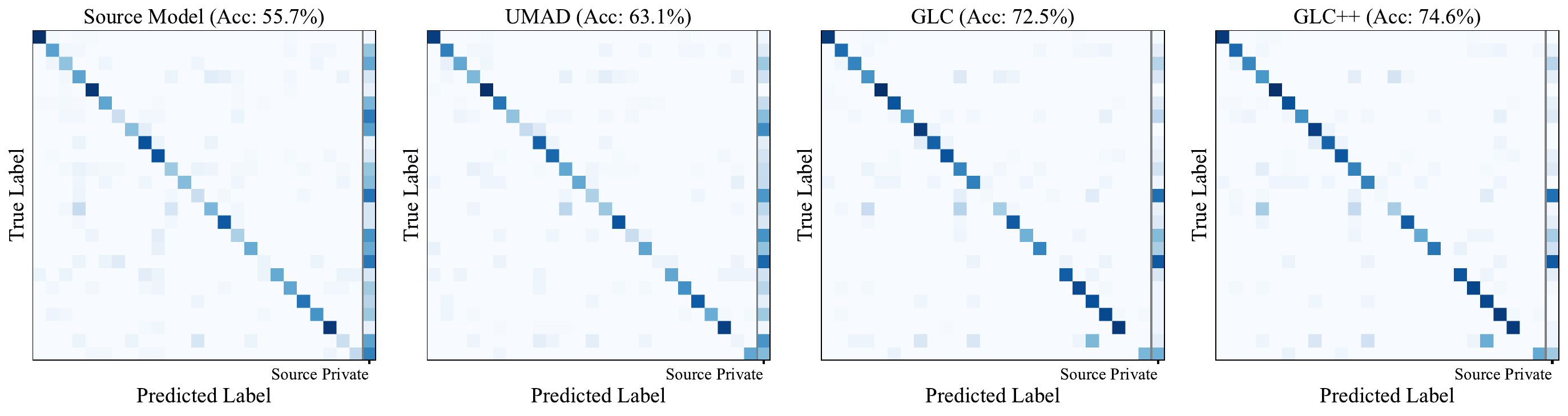}
    \vspace{-0.10in}
    \caption{The confusion matrices for Source Model, UMAD, GLC, and GLC++ in the PDA task Cl$\rightarrow$Ar on the Office-Home dataset. GLC and GLC++ demonstrate less confusion between source common and source private classes.}
    \vspace{-0.15in}
    \label{fig:confusion_mat}
\end{figure*}
\begin{figure*}[ht]
    \centering
    \includegraphics[width=0.99\textwidth]{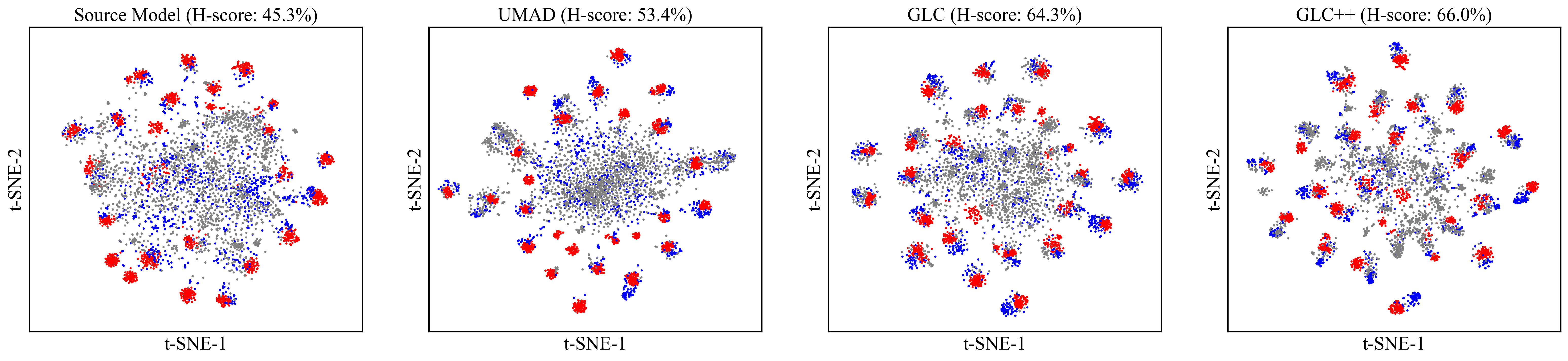}
    \vspace{-0.10in}
    \caption{The t-SNE feature visualizations for Source Model, UMAD, GLC, and GLC++ in the OSDA task Re$\rightarrow$Cl on the Office-Home dataset. {Points in red} denote unavailable source-common data, {points in blue} illustrate target-common data, {points in gray} illustrate target-private data, individually. It is easy to conclude that GLC and GLC++ achieve better feature alignment between the source and target domain, and GLC++ is effective in achieving a clear separation of different clusters among target-private "unknown" data.}
    \vspace{-0.15in}
    \label{fig:tsne}
\end{figure*}
\\
\noindent \textbf{Hyper-parameter Sensitivity:} To better evaluate the effects of different hyper-parameters, we perform a detailed analysis of their sensitivity. {Specifically, we explore the parameter sensitivity of $\eta$, $\rho$, $|L^i|$, and $\gamma$ on Office-31 in OPDA scenarios in Fig. \mbox{\ref{fig:hyper_param}} (a-d), where $\eta$ is varied across the range [0.1, 0.3, 0.5, 1.0, 2.0], $\rho$ spans [1/2, 2/3, 3/4, 4/5, 1.0], $|L^i|$ is in the range of [2, 3, 4, 5, 6], and $\gamma$ with [0.1, 0.3, 0.5, 1.0, 2.0].} Note that $\rho = 1.0$ indicates that we do not introduce the confidence-based source-private suppression mechanism. {It is easy to find that results around the selected parameters $\eta = 0.3$, $\rho = 0.75$, $|L^i|$ are stable, and much better than the source model. This robustness can be attributed to our global one-vs-all clustering strategy, which enables reliable differentiation between "known" source categories and "unknown" target-private categories across various category shift scenarios.}
{In Fig. \mbox{\ref{fig:hyper_param}} (d), we examine the sensitivity of $\gamma$ when we integrate the contrastive affinity learning strategy into UMAD. As demonstrated, the performance remains relatively stable and is not highly sensitive to the selection of $\gamma$.} Additionally, in Fig.~\ref{fig:hyper_param} (e), we further present the H-score concerning different thresholds $\omega$ on Office-Home in the OPDA setting. 
{As stated in Sec. 3.7, we leverage the threshold $\omega$ to distinguish between "known" and "unknown" data. A flatter range of $\omega$ indicates better separation between these categories, as it suggests that the model maintains robustness across a wider threshold range. In contrast, if the reasonable range of $\omega$ is very limited, the model struggles to effectively separate "known" and "unknown" categories, indicating lower confidence in its decision boundary. The results demonstrate that our GLC and GLC++ effectively balance classifying known categories while identifying unknown ones.}
\\
\noindent \textbf{Training Stability}: We investigate the training stability of our models, GLC and GLC++, as illustrated in Fig.~\ref{fig:convergence}, in the OPDA task Cl$\rightarrow$Pr and the OSDA task Re$\rightarrow$Cl on the Office-Home dataset. Both GLC and GLC++ demonstrate stable training convergence, as evidenced by the rapid and consistent increase in the H-score, which begins to stabilize after 10 epochs. However, GLC exhibits limitations in effectively discriminating between different semantic clusters within the "unknown" data category, as indicated by the fluctuating trend in its NCD Acc. In contrast, GLC++ benefits from our contrastive affinity learning strategy, which markedly enhances its ability to identify novel categories within the "unknown" data, consistently outperforming the pre-trained source model. Overall, the training processes for both GLC and GLC++ are stable and exhibit effective performance.\\
\noindent \textbf{Noisy Label Learning}: While covariate and categorical shifts pose significant challenges to model adaptation, real-world scenarios often involve additional complexities, such as noisy or incorrectly labeled source data. To evaluate the robustness of our proposed methods under such conditions, we conduct additional experiments on the Office-31 dataset in both OPDA and OSDA settings. We introduce varying degrees of instance-independent label noise into the source domain, following recent advancements in learning from noisy labels \mbox{\cite{yu2020label,song2022learning,feng2023road}}. As illustrated in Figure \mbox{\ref{fig:src_noise}}, although label noise substantially impairs the performance of the source model, our GLC and GLC++ frameworks consistently demonstrate robust and superior target performance. This resilience is primarily attributed to the interplay among our global one-vs-all clustering strategy, local K-NN consensus clustering mechanism, and contrastive affinity learning. Together, these components mitigate the adverse effects of noisy supervision by enhancing the reliability of pseudo-labeling and fostering robust feature alignment.\\
\noindent \textbf{Varying Unknown Classes:} As the number of "unknown" classes increases, distinguishing between "known" and "unknown" objects becomes more challenging. To assess the robustness of GLC and GLC++, we conduct a comparative analysis with other methods, varying the "unknown" classes in the OPDA task Ar$\rightarrow$Re, and the OSDA task Cl$\rightarrow$Pr on the Office-Home dataset. A notable and consistent observation is that the H-score demonstrates greater stability compared to the NCD Acc. We hypothesize that this discrepancy arises because the H-score metric only requires the model to distinguish "unknown" data from "known" categories. In contrast, the NCD Acc metric necessitates distinguishing between individual "unknown" categories, making the task inherently more challenging as the number of "unknown" categories increases. This challenge is exacerbated by the absence of supervision and the presence of domain and category shifts. The source-available method, OVANet, tends to outperform the source-free methods, likely due to the implicit supervision of source data. {Despite these difficulties, our contrastive affinity learning strategy effectively alleviate this challenge, enabling GLC++ to outperform existing methods in terms of NCD Acc, even as the number of "unknown" categories grows. These results further underscore the robustness and effectiveness of the proposed strategy.}\\
\noindent \textbf{Confusion Matrix:} Fig.~\ref{fig:confusion_mat} visualizes the confusion matrices in the PDA task Cl$\rightarrow$Ar on the Office-Home dataset. The confusion matrices show that the source model and UMAD predominantly misclassify common data, erroneously assigning them to source-private categories. This tendency indicates a challenge in differentiating between common and source-private data in the presence of distribution shifts. In contrast, both GLC and GLC++ exhibit a markedly reduced level of confusion between source common and source private classes. This improvement suggests that our methods are more versatile and effective.\\
\noindent \textbf{Feature Visualization:} Fig.~\ref{fig:tsne} presents the t-SNE visualization of features from the source model, UMAD, GLC and GLC++ in the OSDA task Re$\rightarrow$Cl on the Office-Home dataset. As expected, GLC and GLC++ achieve better feature alignment between the source and target domain. Taking a closer look at the visualization, GLC++ exhibits a significant enhancement over GLC in achieving separation among the categories of target-private "unknown" data, demonstrating the efficacy of our contrastive affinity learning strategy.

\section{Conclusion}
\par In this paper, we have proposed a generic framework for source-free universal domain adaptation termed \emph{Global and Local Clustering (CLC)}. Technically, we have devised an innovative one-vs-all global clustering strategy to realize "unknown" and "known" data separation and introduced a local k-NN clustering strategy to alleviate negative transfer. Compared to existing approaches that require source data or are only applicable to specific category shifts, GLC is appealing by enabling universal model adaptation on the basis of only closed-set source models. We further present a contrastive affinity learning strategy to elevate GLC to GLC++, enhancing the ability of novel category discovery among the target private "unknown" data.  Extensive experiments in partial-set domain adaptation (PDA), open-set domain adaptation (OSDA), open-partial-set domain adaptation (OPDA), and the vanilla closed-set domain adaptation (CLDA) scenarios across several benchmarks have verified the effectiveness and superiority of GLC and GLC++. Remarkably, GLC and GLC++ attain an H-score of 73.2\%/75.3\% in the OPDA task on the VisDA benchmark, which are 16.8\%/18.9\% higher than GATE respectively.
\par \noindent\textbf{Acknowledgment:} 
This work was supported by the National Key Research and Development Program of China (No 2024YFE0211000), the National Natural Science Foundation of China (No. 62372329), the Shanghai Scientific Innovation Foundation (No. 23DZ1203400), the China Postdoctoral Science Foundation (No. GZB20250385, No. 2025M771539), the Tongji-Qomolo Autonomous Driving Commercial Vehicle Joint Lab Project, and the Xiaomi Young Talents Program.

\bibliographystyle{IEEEtran}
\bibliography{main_ref}

\vspace{-0.3in}
\begin{IEEEbiography}[{\includegraphics[width=1in, height=1.25in, clip, keepaspectratio]{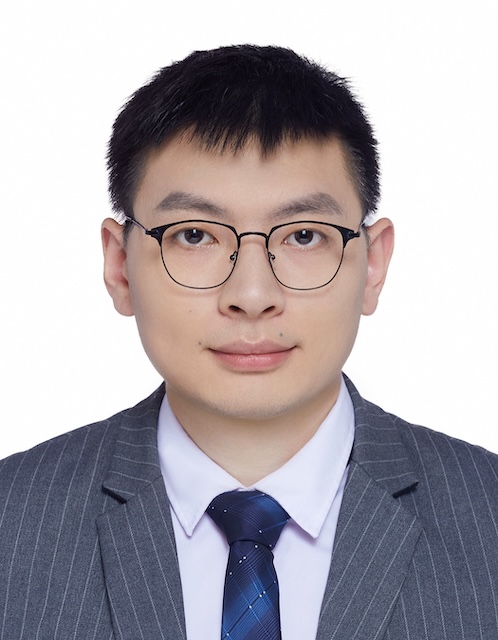}}]{Sanqing Qu} received the B.E. degree and Ph.D. degree in Vehicle Engineering from Tongji University in 2020 and 2024. He is currently a postdoctoral researcher in the School of Computer Science and Technology at Tongji University. His research interests include transfer learning, pattern recognition, and embodied AI.
\end{IEEEbiography}

\vspace{-0.3in}
\begin{IEEEbiography}[{\includegraphics[width=1in, height=1.25in, clip, keepaspectratio]{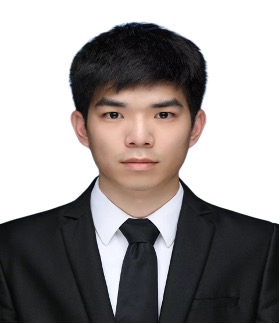}}]{Tianpei Zou} received the B.E. degree in Vehicle Engineering from Hunan University in 2020. He is currently pursuing the Ph.D. degree in vehicle engineering at Tongji University. His research interests include 3D computer vision, deep learning, and transfer learning.
\end{IEEEbiography}

\begin{IEEEbiography}[{\includegraphics[width=1in, height=1.25in, clip, keepaspectratio]{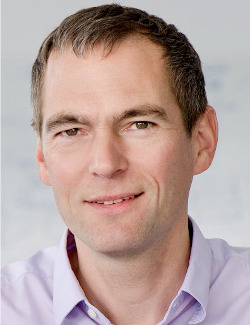}}]{Florian Röhrbein} (Senior Member, IEEE)  received his diploma and Ph.D. degree from the Technical University of Munich (TUM) and the venia legendi for computer science from the University of Bremen. He is a professor with neurorobotics at TU Chemnitz. He is also responsible for the development and implementation of the artificial intelligence (AI) strategy for a world-leading company and is the chief editor of Frontiers in Neurorobotics. He was also the managing director in the Human Brain Project at TUM. He has international experience in various projects on AI, computational neuroscience and brain-inspired cognitive systems.
\end{IEEEbiography}
\vspace{-0.3in}

\begin{IEEEbiography}[{\includegraphics[width=1in, height=1.25in, clip, keepaspectratio]{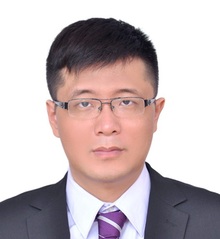}}]{Cewu Lu}  received the PhD degree from the Chinese University of Hong Kong, supervised by Prof. Jiaya Jia. He is a professor with Shanghai Jiao Tong University (SJTU). Before joining SJTU, he was a research fellow with Stanford University, working under Prof. Fei-Fei Li and Prof. Leonidas J. Guibas. He was a research assistant professor with the Hong Kong University of Science and Technology with Prof. Chi Keung Tang. His research interests fall mainly in computer vision, deep learning, and robotics.
\end{IEEEbiography}

\vspace{-0.3in}
\begin{IEEEbiography}[{\includegraphics[width=1in, height=1.25in, clip, keepaspectratio]{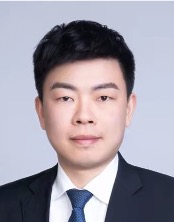}}]{Guang Chen} received the B.S. and M.Eng. degrees in mechanical engineering from Hunan University, China, and the Ph.D. degree from the Faculty of Informatics, Technical University of Munich, Germany, in 2016. He is a professor with Tongji University and a Senior Research Associate (guest) with the Technical University of Munich. He is leading the Robotics and Embodied Artificial Intelligence Laboratory, at Tongji University. His research interests include computer vision and embodied AI. He serves as an associate editor for several international journals.
\end{IEEEbiography}

\vspace{-0.3in}
\begin{IEEEbiography}[{\includegraphics[width=1in, height=1.25in, clip, keepaspectratio]{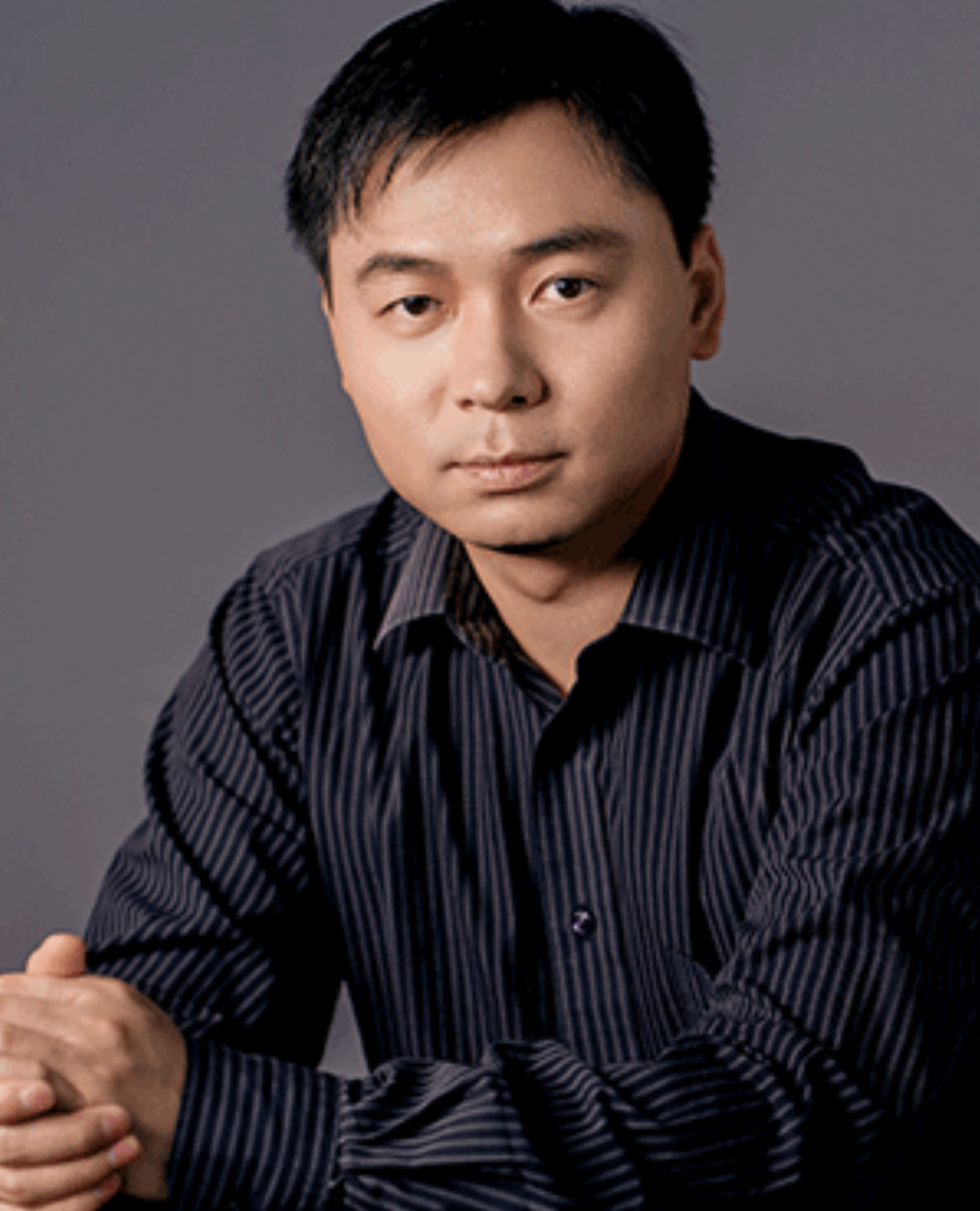}}]{Dacheng Tao} (Fellow, IEEE) is currently a distinguished University professor in the College of Computing \& Data Science with Nanyang Technological University. He mainly applies statistics and mathematics to artificial intelligence and data science, and his research is detailed in one monograph and more than 200 publications in prestigious journals and proceedings, leading conferences, with best paper awards, best student paper awards, and test-of-time awards. He received the 2015 and 2020 Australian Eureka Prize, the 2018 IEEE ICDM Research Contributions Award, and the 2021 IEEE Computer Society McCluskey Technical Achievement Award. He is a Fellow of the Australian Academy of Science, AAAS, ACM and IEEE.
\end{IEEEbiography}

\vspace{-0.3in}
\begin{IEEEbiography}[{\includegraphics[width=1in, height=1.25in, clip, keepaspectratio]{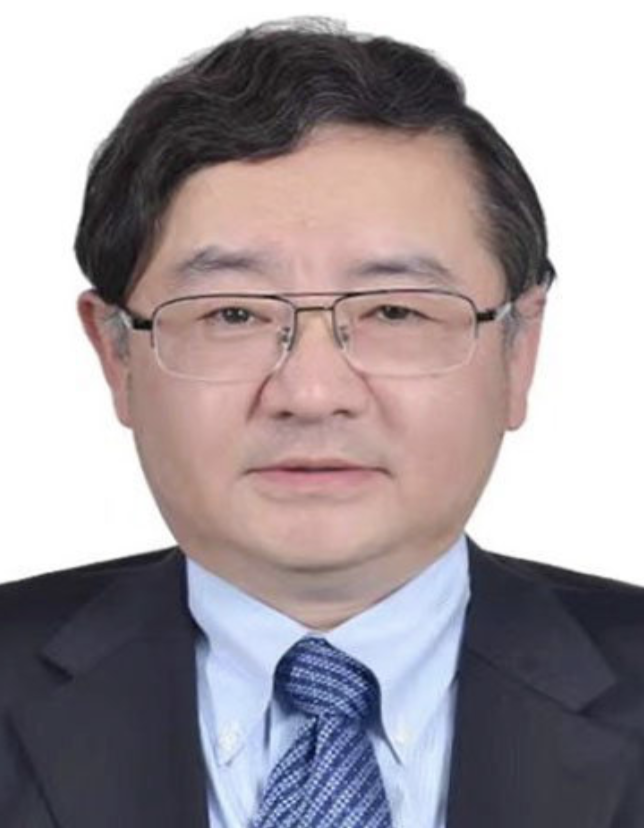}}]{Changjun Jiang} received the Ph.D. degree from the Institute of Automation, Chinese Academy of Sciences, Beijing, China, in 1995. He is currently a professor with the School Computer Science and Technology, Tongji University, Shanghai, China. He has published more than 200 publications. His current research interests include network computing, concurrency theory, the formal verification of software, intelligent systems, and service-oriented computing.
\end{IEEEbiography}

\end{document}